\documentclass[11pt,english]{article}
\usepackage[utf8]{inputenc} 
\usepackage{geometry}
\usepackage{babel}
\usepackage{float}
\usepackage{appendix}
\usepackage{algorithm}
\usepackage{algpseudocode}
\usepackage{amssymb}
\usepackage{booktabs}
\usepackage{multirow}
\usepackage{rotating}
\usepackage{graphicx}
\usepackage{subcaption}
\usepackage{setspace}
\usepackage{comment}
\usepackage{threeparttable}
\usepackage{mathrsfs}
\usepackage{array}
\usepackage{pifont}
\newcommand{\cmark}{\ding{51}}%
\newcommand{\xmark}{\ding{55}}%

\usepackage{caption}
\usepackage[authoryear]{natbib}
\usepackage[unicode=true,pdfusetitle,
 bookmarks=true,bookmarksnumbered=false,bookmarksopen=false,
 breaklinks=false,pdfborder={0 0 0},pdfborderstyle={},backref=false,colorlinks=false]
 {hyperref}

\usepackage{titlesec} 

\usepackage{newtxtext}
\usepackage{newtxmath}
\usepackage{comment}
\usepackage{bbm}

\makeatletter

\hypersetup{colorlinks=true,linkcolor=blue,citecolor=blue,urlcolor=blue}
\usepackage{diagbox}

\usepackage{amsmath}
\DeclareMathOperator*{\argmax}{arg\,max}
\DeclareMathOperator*{\argmin}{arg\,min}

\usepackage{tikz,xcolor}

\makeatother

\begin{document}
\title{Improving Traffic Signal Data Quality for the Waymo Open Motion Dataset}

\author{Xintao Yan\textsuperscript{a}, Erdao Liang\textsuperscript{b}, Jiawei Wang\textsuperscript{a}, Haojie Zhu\textsuperscript{a}, Henry
X. Liu\textsuperscript{a,c,d}\thanks{Corresponding author.}\date{}}

\maketitle

\begin{singlespace}
\textsuperscript{a}\textit{Department of Civil and Environmental
Engineering, University of Michigan, Ann Arbor, MI, USA}

\textsuperscript{b}\textit{Department of Electrical Engineering and Computer Science, University of Michigan, Ann Arbor, MI, USA}

\textsuperscript{c}\textit{Mcity, University of Michigan, Ann Arbor, MI, USA}

\textsuperscript{d}\textit{University of Michigan Transportation
Research Institute, University of Michigan, Ann Arbor, MI, USA}

\end{singlespace}

\section*{Abstract}

Datasets pertaining to autonomous vehicles (AVs) hold significant promise for a range of research fields, including artificial intelligence (AI), autonomous driving, and transportation engineering. Nonetheless, these datasets often encounter challenges related to the states of traffic signals, such as missing or inaccurate data. Such issues can compromise the reliability of the datasets and adversely affect the performance of models developed using them. This research introduces a fully automated approach designed to tackle these issues by utilizing available vehicle trajectory data alongside knowledge from the transportation domain to effectively impute and rectify traffic signal information within the \textsc{Waymo Open Motion Dataset} (WOMD). The proposed method is robust and flexible, capable of handling diverse intersection geometries and traffic signal configurations in real-world scenarios. Comprehensive validations have been conducted on the entire WOMD, focusing on over 360,000 relevant scenarios involving traffic signals, out of a total of 530,000 real-world driving scenarios. In the original dataset, $71.7\%$ traffic signal states are either missing or unknown, all of which were successfully imputed by our proposed method. Furthermore, in the absence of ground-truth signal states, the accuracy of our approach is evaluated based on the rate of red-light violations among vehicle trajectories. Results show that our method reduces the estimated red-light running rate from $15.7\%$ in the original data to $2.9\%$, thereby demonstrating its efficacy in rectifying data inaccuracies. This paper significantly enhances the quality of AV datasets, contributing to the wider AI and AV research communities and benefiting various downstream applications. The code and improved traffic signal data are open-sourced at: \url{https://github.com/michigan-traffic-lab/WOMD-Traffic-Signal-Data-Improvement}.



\hfill\break%
\noindent\textit{Keywords}: Waymo Open Motion Dataset, Traffic Signal, Data Imputation and Correction, Autonomous Vehicles

\newpage

\section{Introduction}

Recent years have witnessed a surge in the release of large-scale datasets pertaining to autonomous vehicles (AVs). Prominent examples include the \textsc{Waymo Open Motion Dataset} (WOMD) \citep{ettinger2021large-womd}, the \textsc{Lyft Level 5 Dataset} \citep{houston2021one-Lyft}, the \textsc{Argoverse Dataset} \citep{wilson2023argoverse-Argoverse2}, and the \textsc{NuScenes Dataset} \citep{caesar2020nuscenes-NuScenes}. These datasets contain hundreds of hours of trajectory data along with contextual information such as high-definition maps and traffic signal states, collected from real-world traffic scenarios. The realistic interactions captured between AVs, human-driven vehicles, pedestrians, and cyclists on the road significantly boost research in various domains, including autonomous driving \citep{chen2023end-data-for-automated-driving} and transportation engineering \citep{li2020trajectory-data-for-transportation-engineering}. Extensive studies have utilized these datasets for trajectory prediction \citep{cui2023gorela-pred, zhou2023query-pred, nayakanti2023wayformer-pred, shi2024mtr++-traj-pred}, AV decision-making \citep{hu2023planning-plan, dauner2023parting-plan, huang2023gameformer-pred-and-plan, cao2023continuous-decision-making}, and human behavior modeling \citep{wen2022characterizing-behavior-model, hu2023autonomous-behavior-model, zhang2023trafficbots-behavior-model, yan2023learning-simulation, huang2024versatile-behavior-model, jiao2024beyond-behavior-model}. These datasets have empowered researchers to tackle fundamental challenges in these fields, establishing important benchmarks \citep{montali2024waymo-WOSAC-benchmark, waymo-motion-prediction-challenge-benchmark, feng2024unitraj-benchmark, dauner2024navsim-benchmark} for future research.

Despite their immense potential, these datasets often encounter challenges related to traffic signal information. Missing values and data errors in traffic signal states can undermine the reliability of these datasets and substantially degrade the performance of models trained on them. These issues stem primarily from the data collection mechanism: AV sensors (\emph{e.g.}, cameras) capture the traffic lights that are applied to their approach only, leading to missing data for other approaches. Additionally, computer vision techniques for camera data processing can introduce errors, resulting in inaccurate traffic signal states. Given the critical role of traffic signal information in various applications, ranging from prediction and decision-making \citep{li2023towards-survey} to simulation \citep{chen2024data-survey}, ensuring the quality of this data is paramount. 

The issue of data noise and inaccuracy in public datasets is not unprecedented. For example, extensive efforts have been made to improve the widely used NGSIM dataset \citep{punzo2011assessment-improve-NGSIM, montanino2015trajectory-improve-NGSIM, coifman2017critical-improve-NGSIM} in the transportation engineering domain. Recently, studies aimed at enhancing the trajectory data quality of the WOMD have emerged \citep{hu2022processing-improve-WOMD-traj}. However, to the best of our knowledge, our study is the first to address the challenge of traffic signal information in AV datasets. There are several fundamental differences between existing Signal Phase and Timing (SPaT) estimation studies and our research, with respect to problem settings and data availability. Existing SPaT estimation studies aim to estimate complete signal timing plans \citep{yu2016learning-SPaT-estimation, du2019signal-SPaT-estimation, mahmud2023use-SPaT-estimation}, including cycle length, green splits, and time-of-day, by leveraging partially observed floating car data. Given that only a small percentage of vehicles are observable (\emph{e.g.}, a 5\% penetration rate in real-world practice), these studies require the accumulation of large amounts of data across multiple days to infer the SPaT for each intersection. In contrast, our study focuses on data segments with full observation of all vehicle trajectories, but for only very short time intervals (9-20 seconds) at each intersection. Our goal is to estimate traffic signal states during the data intervals. Due to these differences, the methodologies developed in existing SPaT estimation studies cannot be directly applied to our problem.

\begin{figure}[t!]
    \centering
        \includegraphics[width=1\textwidth]{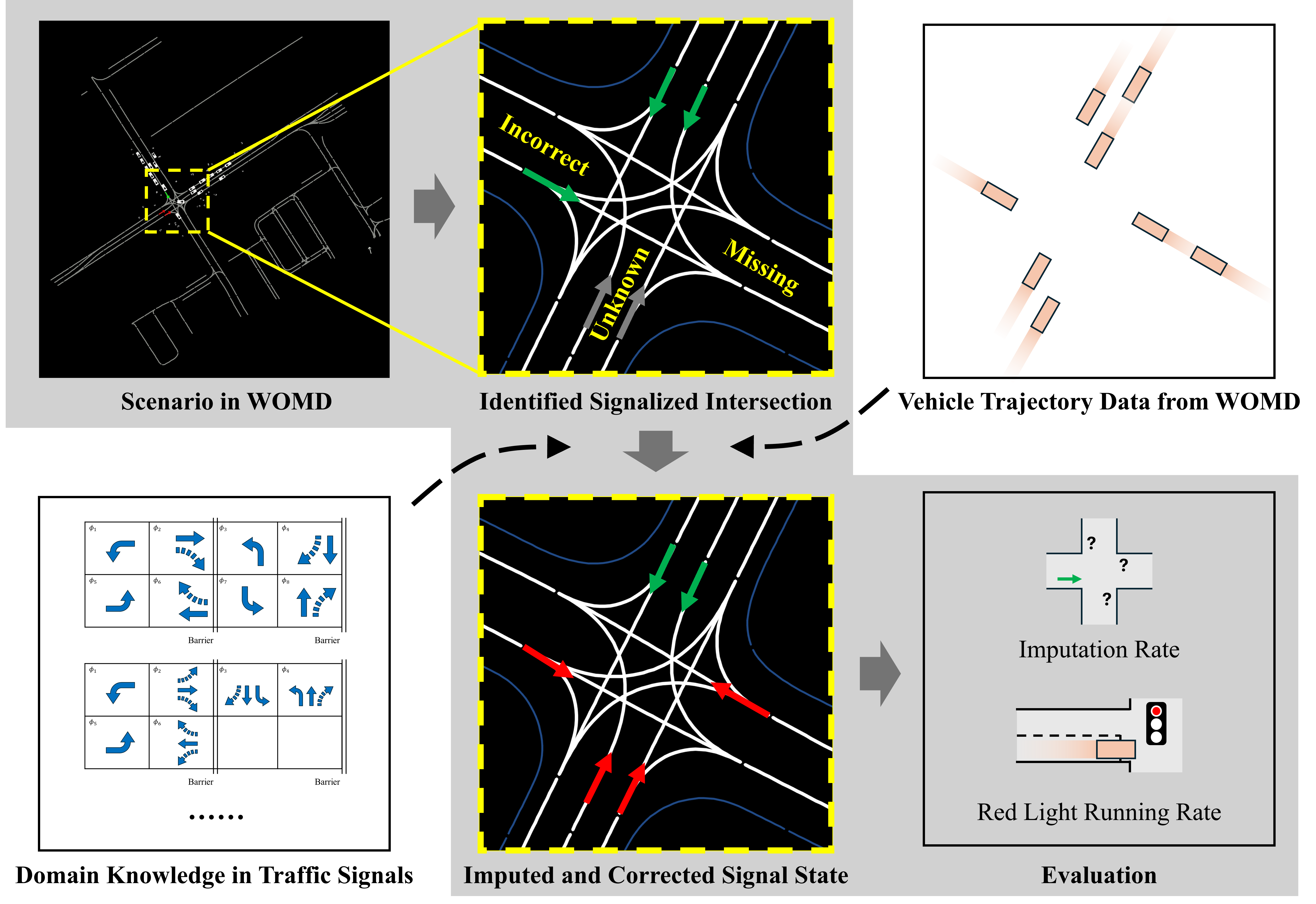}  
    \caption{Method schematic for imputing and correcting traffic signal states in the WOMD.}
    \label{fig:main}
\end{figure}

In this study, we propose a fully automated method that leverages map data, trajectory data, and transportation domain knowledge to impute missing traffic signal data and correct erroneous entries for AV datasets. Specifically, WOMD is utilized as an example to demonstrate the performance, as it is one of the most representative and widely used AV datasets. Figure~\ref{fig:main} demonstrates the schematic for our method. Human driving behavior, as reflected in trajectory data, can provide valuable insights into traffic signal states. For instance, if a vehicle is approaching a stop bar at high velocity without deceleration, it is most likely that the traffic signal light is green. Accordingly, we can estimate the corresponding traffic signal states by analyzing trajectory data. Furthermore, most signalized intersections in the United States follow the National Electrical Manufacturers Association (NEMA) standards with Ring-and-Barrier diagram structure \citep{urbanik2015signal-signal-mannual}. This domain knowledge can guide the estimation process, especially for data segments with limited vehicle presence.  To validate the proposed method's performance, we will evaluate the imputation rate and the red-light running rate across a large number of scenarios, demonstrating its effectiveness.

This study is motivated by the need to enhance the data quality of the WOMD, thereby benefiting all downstream applications that utilize this dataset across various communities. Our method is both robust and flexible, capable of adapting to diverse intersection geometries and traffic signal configurations with high accuracy. This effort highlights the contributions of transportation researchers to the broader artificial intelligence (AI) and AV communities. The remainder of this paper is organized as follows: Section 2 provides the problem statement, Section 3 introduces the methodology, Section 4 presents results, and Section 5 summarizes the main conclusions and discusses future directions.

\begin{figure}[ht!]
    \centering
    \includegraphics[width=1\linewidth]{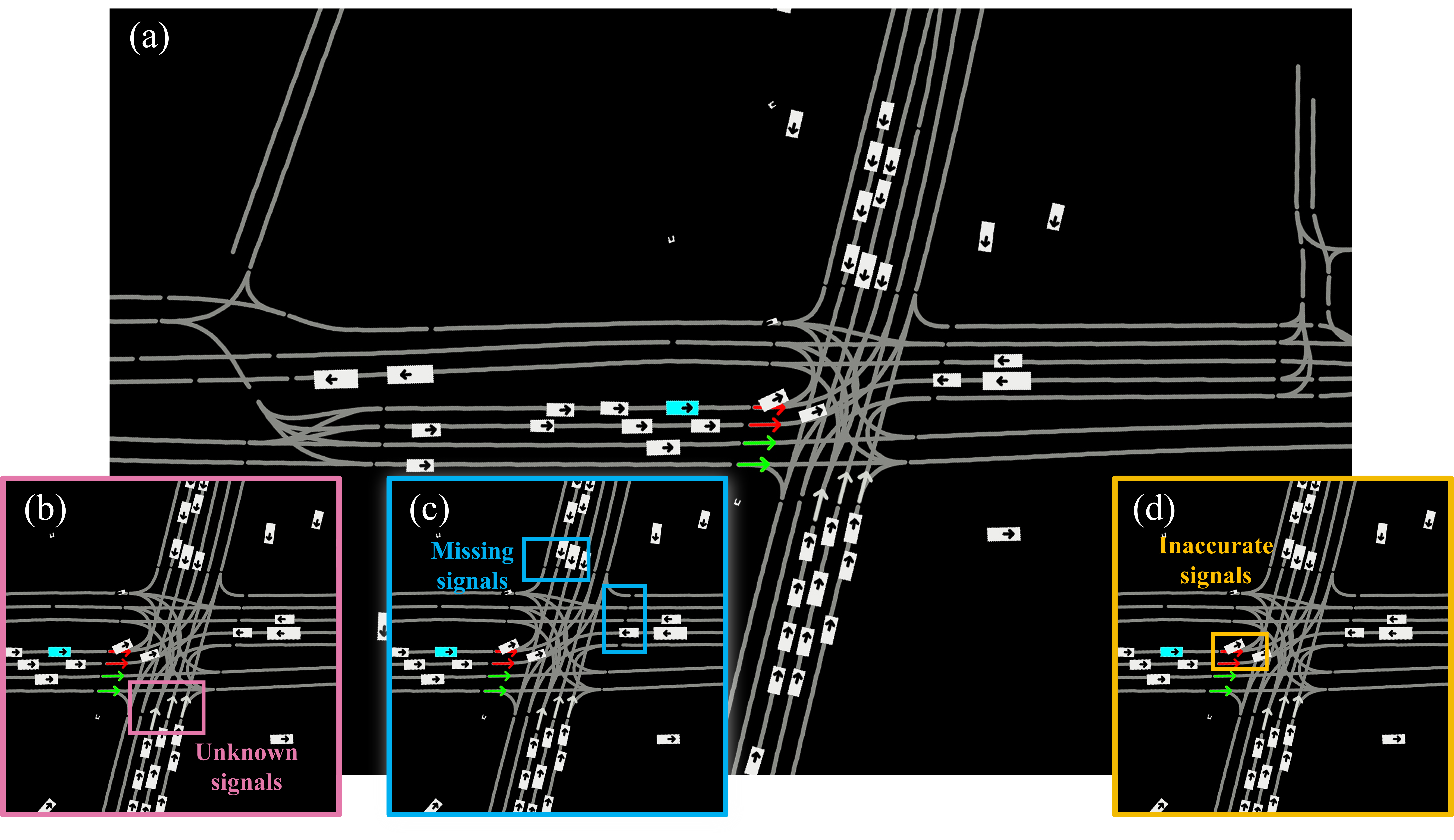}
    \vspace{2mm}
    \caption{Demonstration of the WOMD and the traffic signal data issues. (a) A snapshot of a scenario in the WOMD, (b) Unknown traffic signal information, (c) Missing traffic signal information, (d) Inaccurate traffic signal information.}
    \label{fig:waymo-data-and-issues-demo}
\end{figure}

\section{Problem Statement}

In this section, we first introduce the issues inherent in the original traffic signal information within the WOMD. We then proceed to formulate the main problem to be addressed.

\subsection{Traffic Signal Issues in the Waymo Open Motion Dataset}
The WOMD \citep{ettinger2021large-womd}, released by Waymo LLC, is one of the most widely used datasets in autonomous driving research and supports key tasks such as trajectory prediction \citep{shi2024mtr++-traj-pred, sun2025impact}, motion planning \citep{huang2023gameformer-pred-and-plan}, and simulation modeling \citep{zhang2024closed}. The dataset was collected across six cities in the United States, covering over 1,750 kilometers of roadways. It contains more than 530,000 scenarios, where each scenario is a 9-second data segment sampled at 10 Hz, encompassing trajectory data for all surrounding agents (vehicles, pedestrians, cyclists) near the Waymo self-driving car (SDC). In addition to the trajectory data, each scenario includes static map information such as road centerlines, road edges, and stop sign locations for the covered area. It also features traffic signal data, detailing the position of each signal, the lanes it controls, and the state of the traffic light (\emph{e.g.}, green, yellow, red) at each moment. Among all scenarios, over 360,000 involve signalized intersections, accounting for approximately 68\% of the dataset. For a detailed description of the WOMD, please refer to here\footnote{\url{https://waymo.com/open/data/motion/}}. Figure~\ref{fig:waymo-data-and-issues-demo}a shows a snapshot of a WOMD scenario, with filled bounding boxes representing the agents, the SDC highlighted in cyan, gray lines indicating road centerlines, and colored arrows denoting traffic signal information. Note that for clarity, we only display lane centerlines from the static map information, omitting other details such as road edges and crosswalks.

As shown in Figures~\ref{fig:waymo-data-and-issues-demo}b-d, there are three types of issues for the traffic signal information in the WOMD, detailed as follows: 

\begin{itemize}
\item \textbf{Unknown Traffic Signal States:} As illustrated in Figure~\ref{fig:waymo-data-and-issues-demo}b, some locations in the WOMD have traffic signals, but their states are not provided. This means the dataset indicates the presence of a traffic signal without specifying its current state.
\item \textbf{Missing Traffic Signal Information:} As illustrated in Figure~\ref{fig:waymo-data-and-issues-demo}c, some positions lack traffic signal information completely. In these cases, the raw dataset does not provide any details about the traffic signals at these locations. To address this issue, we need to first identify positions with missing traffic signal information and then estimate their states.

\item \textbf{Inaccurate Traffic Signal States:} As illustrated in Figure~\ref{fig:waymo-data-and-issues-demo}d, the given traffic signal state could have errors. The original data show that the left turn traffic signal is red. However, vehicle trajectory data from before and after this moment show that multiple vehicles began moving from a stationary position and crossed the stop line while the light was red, suggesting potential errors in the traffic signal data.

\end{itemize}
The first two issues are identified as the incompleteness of traffic signal information, necessitating data imputation. The third issue is categorized as inaccuracy, which requires data correction. This study aims to develop a robust method to automatically address all three issues to enhance the WOMD data quality and validate its performance on the entire dataset.

\subsection{Problem Formulation}

In this paper, we employ map data, vehicle trajectory data, and transportation domain knowledge to impute and correct traffic signal state in WOMD, as shown in Figure~\ref{fig:main}. The specific formulation is presented as follows.

\begin{figure}[t!]
    \centering
    \includegraphics[width=1\linewidth]{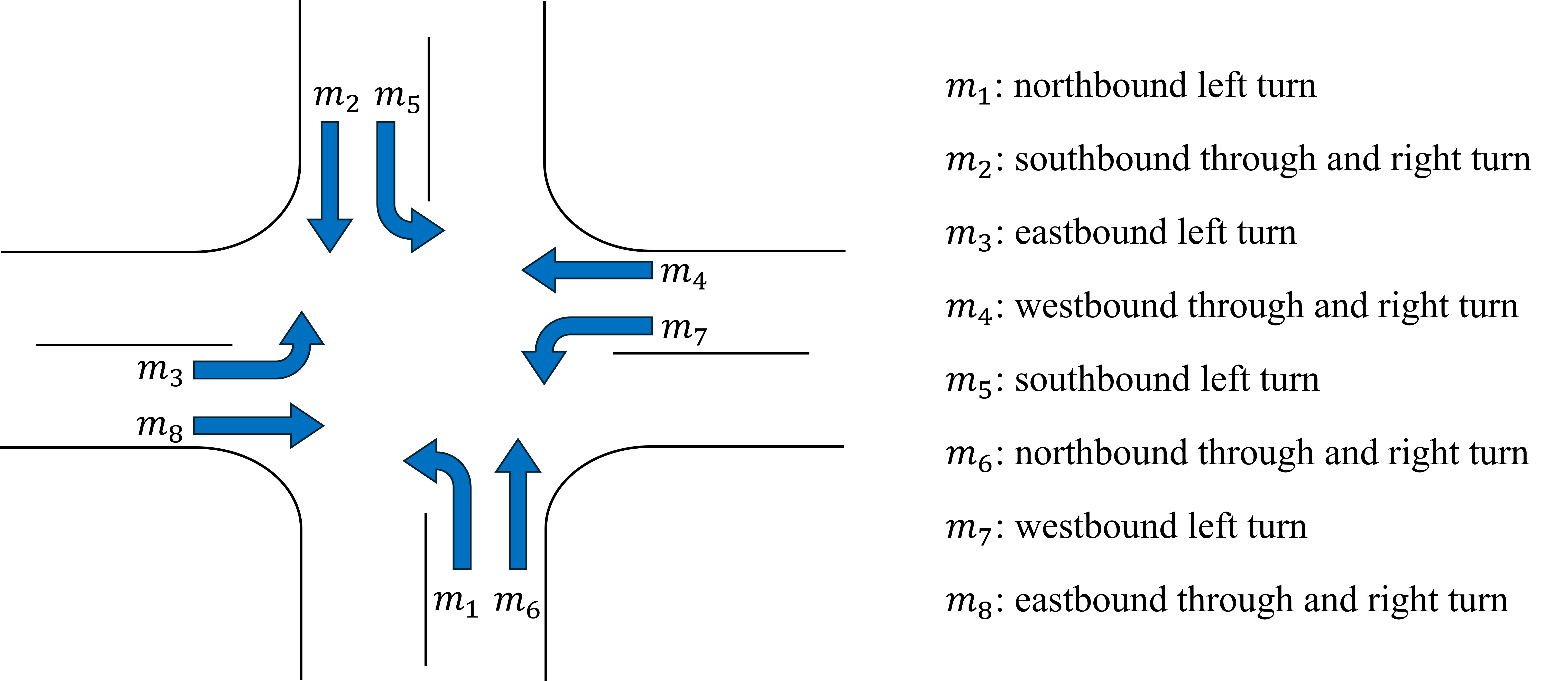}
    \vspace{1.5mm}
    \caption{Demonstration of movements in a typical four-way intersection.}
    \label{fig:movement-intro}
\end{figure}

For a signalized intersection, we assume that it has $N$ movements, denoted as $m_1,\ldots,m_N$. Figure~\ref{fig:movement-intro} illustrates a typical four-way intersection using the movement numbering convention from the Highway Capacity Manual \citep{manual2000highway-HCM}. Right-turn movements are not considered individually because right-turn and through movements from the same approach generally occur simultaneously. It is important to note that a single movement may involve multiple lanes (\emph{e.g.}, a through movement), and a single lane can accommodate multiple movements (\emph{e.g.}, through and right-turn movements). The mapping relationship between movements and road lanes can be derived from the intersection geometry information.

In the WOMD, certain movements at a signalized intersection are explicitly labeled with a traffic light state at each time $t$ ($t=1,2,\ldots,T$, where $T\in \mathbb{N}$ denotes the total number of discrete time instances in this scenario), while other movements are either unknown or missing. We define the traffic light state set as $\Omega=\{G, Y, R, \varnothing\}$, where $G$, $Y$ and $R$ represent the green, yellow, and red lights, respectively, and $\varnothing$ denotes an unknown state (with missing cases also considered as unknown). Thus, in the original WOMD dataset, the state of each movement belongs to the set $\Omega$, \emph{i.e.}, $m^{\mathrm{raw}}_i(t)\in \Omega$, for $t=1,2,\ldots,T$ and $i=1,2,\ldots,N$. The traffic signal state obtained from the original dataset at time $t$ is represented as follows:
\begin{equation}
    x^{\mathrm{raw}}(t)=\begin{bmatrix}
m^{\mathrm{raw}}_1(t),m^{\mathrm{raw}}_2(t),\ldots,m^{\mathrm{raw}}_N(t)\end{bmatrix}.
\end{equation}
Then, the signal state sequence for the scenario is denoted as
\begin{equation}
    \textbf{x}^{\mathrm{raw}} = \begin{bmatrix}
        x^{\mathrm{raw}}(1), x^{\mathrm{raw}}(2), \ldots, x^{\mathrm{raw}}(T)\end{bmatrix}.
\end{equation}

As demonstrated in the previous subsection, every state $x^{\mathrm{raw}}(t)$ in the sequence might suffer from issues of incompleteness and incorrectness. Our objective is to derive a new sequence of signal states, denoted as
\begin{equation}
    \textbf{x}^{*} = \begin{bmatrix}
        x^{*}(1), x^{*}(2), \ldots, x^{*}(T)\end{bmatrix},
\end{equation}
where
\begin{equation}
    x^{*}(t)=\begin{bmatrix}
m^{*}_1(t),m^{*}_2(t),\ldots,m^{*}_N(t)\end{bmatrix}
\end{equation}
is the new traffic signal state at time $t$. $x^{*}(t)$ will be derived based on the raw signal state $x^{\mathrm{raw}}(t)$, the vehicle trajectory data, and domain knowledge in traffic signal timing. Particularly, we seek to find $x^{*}(t)$ that satisfies the following properties:
\begin{itemize}
    \item $x^{*}(t)$ has no unknown states, \emph{i.e.}, all traffic signal states are assigned with an explicit state $G$, $Y$, or $R$, instead of $\varnothing$.
    \item $x^{*}(t)$ is close to the original state $x^{\mathrm{raw}}(t)$;    
    \item $x^{*}(t)$ adheres to the ring-and-barrier structure common in traffic signal timing practice, ensuring no conflicting movements are set to green simultaneously;
    \item Consistency is achieved between vehicle trajectory and signal state, \emph{i.e.}, most vehicles do not run a red light.
\end{itemize}

\section{Methodology}


The framework of our proposed method is illustrated in Figure~\ref{fig:main}. Our approach begins with the original data from the WOMD. We first identify signalized intersections in each scenario by utilizing the map data included in the dataset. Next, we use vehicle trajectory data and signal timing domain knowledge to derive the imputed and corrected signal states. This section provides a detailed explanation of these processes.

\subsection{Identification of Signalized Intersections}

\begin{figure}[t!]
    \centering
    \includegraphics[width=1\textwidth]{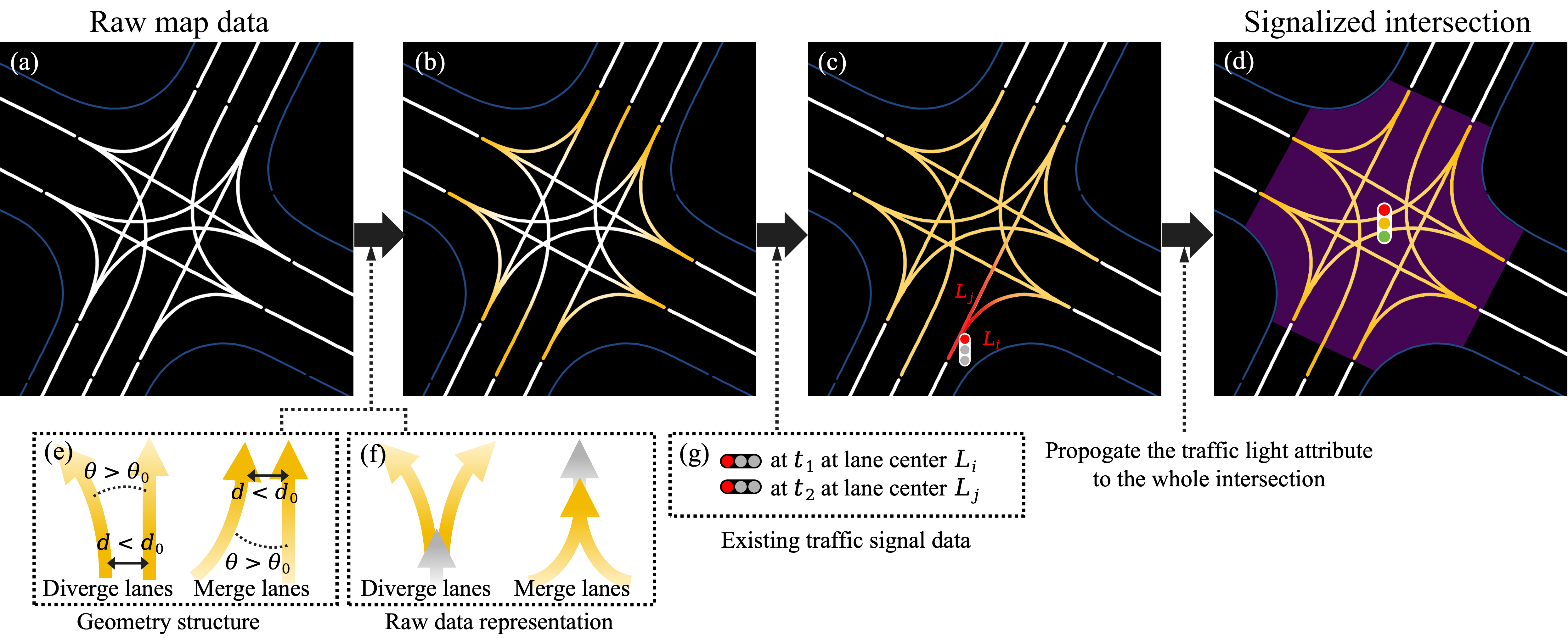} 
    \vspace{2mm}
    \caption{The process of identifying a signalized intersection.}
    \label{fig:identify_tl_intersection}
\end{figure}


In the WOMD, each road centerline is represented by a series of connected lane segments. As shown in Figure~\ref{fig:waymo-data-and-issues-demo}a, each gray line represents a lane segment. Notably, intersections are not explicitly represented in the dataset, meaning the data does not directly indicate the locations of intersections or the lane segments within them. Consequently, the first step in our methodology is to identify intersections in the WOMD using the available lane information.

Special patterns of lane segments can be observed at intersections, as illustrated in Figure \ref{fig:identify_tl_intersection}a and \ref{fig:identify_tl_intersection}b. When entering an intersection, the lane segments diverge from the same starting point and end in different ending directions. Conversely, when leaving the intersection, lane segments converge from different starting points and merge in the same ending direction. These patterns could allow one to group different lanes into lane pairs, including the diverge lane pair or the merge lane pairs.

Along this direction, we employ two methods to identify the lane pairs: 1) Geometric structure (Figure \ref{fig:identify_tl_intersection}e). If two lanes' starting points are close to each other, their ending points are far apart, and their angle exceeds a certain threshold, they are considered a pair of diverge lanes. Similarly, merge lanes are identified if the opposite criteria hold. 2) Raw map data representation (Figure \ref{fig:identify_tl_intersection}f). WOMD map data provides the information of the entry lane and the exit lane for each lane segment. Accordingly, if two lanes share the same entry lane, they are classified as diverge lanes; merge lanes are identified if they share the same exit lane.

After we pair the lanes as diverge or merge, we group them into lane sets through an Union-Find Algorithm~\citep{wikipedia_disjoint_set}. After grouping, we assume that each lane set represents an intersection. To determine which intersections are signalized, we check if any lane in a lane set is associated with traffic signal records from the WOMD. If a lane has an associated signal state, the entire lane set is considered a signalized intersection. This is illustrated in Figure \ref{fig:identify_tl_intersection}c and \ref{fig:identify_tl_intersection}g. Meanwhile, we identify the stopping lines of each intersection, which consist of the starting points of all the lanes in a lane set. In this manner, the boundary of each intersection is also determined. This completes the identification of signalized intersections in the dataset, shown in the purple region in Figure~\ref{fig:identify_tl_intersection}d, providing a foundation for the subsequent steps in our methodology.

Once the signalized intersections are identified, the lane structure approaching each intersection is also determined. This information allows us to pinpoint the various movements within the intersection and obtain the mapping relationship between movements and the corresponding road lanes.

\subsection{Imputation and Correction of Traffic Signal States}

With signalized intersections identified in the previous step, this section introduces how we integrate the original traffic signal data, vehicle trajectory information, as well as the ring-and-barrier diagram structure, to construct a final sequence of signal states $\textbf{x}^* = \begin{bmatrix}
    x^*(1), \ldots, x^*(T)
\end{bmatrix}$ for each intersection.

\subsubsection{Estimating Traffic Signal State via Trajectory Data}
\vspace{0.2em}

The traffic signal state is estimated by leveraging vehicle trajectories that are on this movement. For all the vehicles in the movements $i$, denote the set of their indices as $\mathcal{K}_i$. Note that we do not include right-turning vehicles in the vehicle set, as right turns are permitted during both red and green lights in many intersections, which might confuse the estimation process.

For each vehicle $k$ at time $t$ approaching the intersection, we have its distance to the stopping line $d_k(t)$, its acceleration $a_k(t)$, and its velocity $v_k(t)$. Based on this information, we define two vehicle confidence values $f(d_k(t),a_k(t))$ and $g(d_k(t),v_k(t))$ based on acceleration and velocity, which are calculated for each vehicle at each moment. For the complete form of $f$ and $g$ functions, please refer to the Appendix~\ref{sec:appendix_A}. These vehicle confidence values act as measures of how strongly the trajectory information of this vehicle evidences the presence of a red or green signal. The larger the vehicle confidence value, the higher the confidence we have in this vehicle's information on determining the signal states.

Next, we calculate the weighted mean acceleration $\bar{a}_i(t)$ of all vehicles for this movement over the past and future time horizon $\left[t-\Delta t, t+\Delta t\right]$ ($\Delta t \geq 1$).

\begin{equation}
\bar{a}_i(t) = \frac{\sum_{k\in \mathcal{K}_i}\max_{\tau \in [t-\Delta t, t+\Delta t]} f(d_k(\tau), a_k(\tau))\bar{a}_{ik}}{\sum_{k \in \mathcal{K}_i} \max_{\tau \in [t-\Delta t, t+\Delta t]} f(d_k(\tau), a_k(\tau))},
\label{eqn:mean_acceleration}
\end{equation}
where
\begin{equation}
\bar{a}_{ik} = \frac{\sum_{\tau=t-\Delta t}^{t+\Delta t} f(d_k(\tau),a_k(\tau))a_k(\tau)}{\sum_{\tau=t-\Delta t}^{t+\Delta t}f(d_k(\tau), a_k(\tau))}
\end{equation}
is the weighted mean acceleration of a single vehicle over the time horizon. Similarly,  the weighted mean velocity $\bar{v}_i(t)$ is calculated as follows
\begin{equation}
\bar{v}_i(t) = \frac{\sum_{k\in \mathcal{K}_i}g_k^{\max} = \max_{\tau \in [t-\Delta t, t+\Delta t]} g(d_k(\tau), v_k(\tau))\bar{v}_{ik}}{\sum_{k \in \mathcal{K}_i} g_k^{\max} = \max_{\tau \in [t-\Delta t, t+\Delta t]} g(d_k(\tau), v_k(\tau))},
\label{eqn:mean_velocity}
\end{equation}
where
\begin{equation}
\bar{v}_{ik} = \frac{\sum_{\tau=t-\Delta t}^{t+\Delta t} g(d_k(\tau),v_k(\tau))v_k(\tau)}{\sum_{\tau=t-\Delta t}^{t+\Delta t}g(d_k(\tau), v_k(\tau))}
\end{equation}
is the weighted mean velocity of a single vehicle among the time horizon.

\begin{figure}[b]
    \centering
    \includegraphics[width=0.8\textwidth]{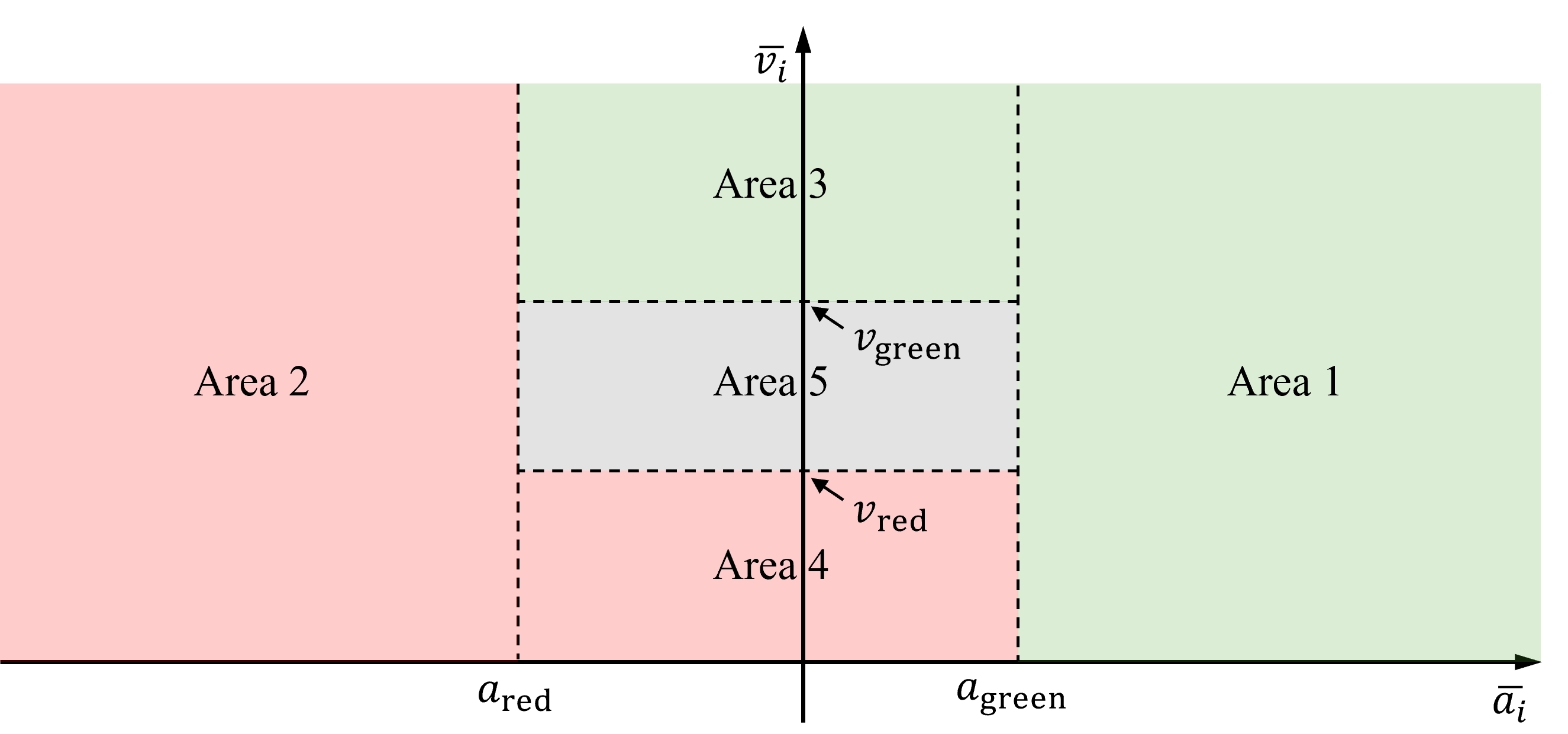} 
    \caption{Area for estimating traffic signal state.}
    \label{fig:rule-estimating}
\end{figure}

Based on this mean acceleration and mean velocity, we derive the estimated movement state $m_i^\mathrm{est}(t)$ following the logic shown in Figure~\ref{fig:rule-estimating} and Table~\ref{table:rule-estimating}, where time step $t$ is neglected for concise expression. Each area represents a specific condition, and the color denotes the estimation results.

\begin{table}[t!]
    \centering
    \begin{threeparttable}
\setlength{\tabcolsep}{20pt}
    \begin{tabular}{ccc}
        \toprule
         & $m_i^\mathrm{est}$ & $c_i$ \\
        \midrule
        Area 1 & $G$ & $\sum_{k\in \mathcal{K}_i} \max_{\tau \in [t-\Delta t, t+\Delta t]} f(d_k(\tau), a_k(\tau))$ \\
        Area 2 & $R$ & $\sum_{k\in \mathcal{K}_i} \max_{\tau \in [t-\Delta t, t+\Delta t]} f(d_k(\tau), a_k(\tau))$\\
        Area 3 & $G$ & $\sum_{k\in \mathcal{K}_i} \max_{\tau \in [t-\Delta t, t+\Delta t]} g(d_k(\tau), v_k(\tau))$ \\
        Area 4 & $R$ & $\sum_{k\in \mathcal{K}_i} \max_{\tau \in [t-\Delta t, t+\Delta t]} g(d_k(\tau), v_k(\tau))$ \\
        Area 5 & $\varnothing$ & 0 \\
        \bottomrule
    \end{tabular}
    \caption{Logic for estimating traffic signal state.}
    \label{table:rule-estimating}
    \end{threeparttable}
\end{table}

In Table~\ref{table:rule-estimating}, we introduce a movement confidence value $c_i(t)$ alongside the estimated signal state $m_i^{\mathrm{est}}(t)$. The movement confidence is calculated by summing the maximum vehicle confidence values of all vehicles within the time horizon. This metric quantifies the reliability of our estimation for the signal state. A high $c_i(t)$ suggests that the vehicle trajectory data provides strong evidence supporting the estimated signal state $m_i^{\mathrm{est}}(t)$, while a low value indicates insufficient or inconclusive data. Consequently, if a discrepancy is detected between the original traffic light state and the estimated one (\emph{i.e.}, $m^{\mathrm{raw}}(t) \neq m^{\mathrm{est}}(t)$, suggesting potential errors in the original data), the estimated state will be adopted if the movement confidence value is sufficiently high.

With $m_i^{\mathrm{est}}(t)$ and $c_i(t)$ for every movement $i$, we have an estimated state $x^\mathrm{est}(t)=\begin{bmatrix}
    m^\mathrm{est}_1(t),\ldots,m^\mathrm{est}_N(t)
\end{bmatrix}$ at a particular time instance $t$, as well as its corresponding confidence value $c(t)=\begin{bmatrix}
    c_1(t),\ldots,c_N(t)
\end{bmatrix}$.

\subsubsection{Imputing and Correcting Signal State}

Now we have the original signal state $x^\mathrm{raw}(t)$ from the dataset, the estimated signal state $x^\mathrm{est}(t)$ from trajectory data, as well as its corresponding confidence value $c(t)$. We proceed to show the imputation and correction method to get the final signal state $x^*(t)$ for time $t$.

Define $x^\mathrm{imp}(t) = \begin{bmatrix}
    m^\mathrm{imp}_1(t), \ldots, m^\mathrm{imp}_N(t)
\end{bmatrix}$ as the imputed signal states derived from $x^\mathrm{raw}(t)$ and $x^\mathrm{est}(t)$. The logic used to determine its value is detailed in Table~\ref{table:rule-imputing}, where the time step $t$ is omitted for simplicity. For each movement $i$, $m^\mathrm{imp}(t)$ is chosen between the original signal state $m^\mathrm{raw}(t)$ and the estimated signal state $m^\mathrm{est}(t)$. If the original movement state is unavailable, the estimated value is used for imputation. If the estimated movement confidence value exceeds a threshold $\theta$ and the state differs from the original data, the original data is corrected with the estimated value. Other specific scenarios and conditions are outlined in the table.

\begin{table}[h!]
    \centering
    \begin{threeparttable}
\setlength{\tabcolsep}{20pt}
    \begin{tabular}{cccc|cc}
        \toprule
        \multicolumn{4}{c|}{\textbf{Conditions}} & \multicolumn{2}{c}{\textbf{Results}} \\
        $m_i^\mathrm{raw}=\varnothing$ & $m_i^\mathrm{est}=\varnothing$ & $m_i^\mathrm{raw}=m_i^\mathrm{est}$ & $c_i\geq \theta $ & $m_i^\mathrm{imp}$ & $w_i$ \\
        \midrule
        \cmark & \cmark & / & / & $\varnothing$ & $0$ \\
        \cmark & \xmark & / & / & $m_i^\mathrm{est}$ & $c_i$\\
        \xmark & \cmark & / & / & $m_i^\mathrm{raw}$ & $w_{\mathrm{small}}$ \\
        \xmark & \xmark & \cmark & / & $m_i^\mathrm{raw}$ & $w_{\mathrm{big}}$ \\
        \xmark & \xmark & \xmark & \cmark & $m_i^\mathrm{est}$ & $c_i$ \\
        \xmark & \xmark & \xmark & \xmark &  $m_i^\mathrm{raw}$ & 0 \\
        \bottomrule
    \end{tabular}
    \caption{Logic for imputing and correcting traffic signal state.}
    \label{table:rule-imputing}
    \end{threeparttable}
\end{table}

However, there may still be empty values in $x^\mathrm{imp}(t)$, as illustrated in the first case in Table~\ref{table:rule-imputing}. Furthermore, conflicts between movements in \(x^\mathrm{imp}(t)\) can emerge. For example, multiple conflicting movements might be assigned green lights, which violates the ring-and-barrier diagram and would not occur in real-world scenarios. To resolve these issues, we acknowledge that signal state \(x(t)\) in real-world applications is restricted to a limited set of feasible configurations. According to the ring-and-barrier diagram, only certain combinations of movements can simultaneously display a green light at any given time instance. Using the standard four-way traffic signal shown in Figure~\ref{fig:movement-intro} as an example, Table~\ref{table:possible-choices} enumerates all ten feasible configurations, where \(G\) indicates a green light and blank cells indicate a red light for each movement.

\begin{table}[h!]
    \centering
    \begin{threeparttable}
\setlength{\tabcolsep}{15pt}
    \begin{tabular}{ccccccccc}
        \toprule
        & $m_1$ & $m_2$ &$m_3$ &$m_4$ &$m_5$ & $m_6$ &$m_7$ &$m_8$ 
         \\
        \midrule
        1& $G$ & & & &$G$ & & & \\
        2& $G$& & & & &$G$ & & \\
        3& & $G$& & & $G$& & & \\
        4& & $G$& & & & $G$& & \\
        5& & & $G$& & & & $G$& \\
        6& & & $G$& & & & & $G$\\
        7& & & & $G$& & & $G$& \\
        8& & & & $G$& & & &$G$ \\
        9& $G$ & $G$ & & & $G$ & $G$ & & \\
        10& & & $G$ & $G$ & & & $G$ &$G$ \\        
        
        \bottomrule
    \end{tabular}
    \caption{Feasible green light configurations for a four-way intersection, adhering to the ring-and-barrier diagram. The meaning of the movements corresponding to those in Figure~\ref{fig:movement-intro}.}    
    \label{table:possible-choices}
    \end{threeparttable}
\end{table}

Given this real-world implementation consideration, we have a practical constraint of $x(t)\in \mathcal{X}$, where the feasible set $\mathcal{X}$ contains all the possible choices (\emph{e.g.}, the ten choices in Table~\ref{table:possible-choices} for a four-way intersection). 
We then need to find the closest match in $\mathcal{X}$ to $x^\mathrm{imp}(t)$, which will be designated as our final signal state $x^*(t)$.

To quantify the similarity of $x^\mathrm{imp}(t)$ to any feasible choice $x^{\mathrm{feas}}(t)$, we introduce a weight coefficient $w_i(t)$ for each movement state in Table \ref{table:rule-imputing}, forming $w(t)=\begin{bmatrix}w_1(t), \cdots, w_N(t)\end{bmatrix}$. Intuitively, $w_i(t)$ measures the degree of certainty regarding the imputed state $m^{\mathrm{imp}}_i(t)$: the higher our confidence that $m^{\mathrm{imp}}_i(t)$ is correct, the larger the $w_i(t)$ value. The detailed logic is as follows:
\begin{itemize}
    \item When the estimated movement state matches the original data state, \emph{i.e.}, $m_i^\mathrm{imp}(t)=m_i^\mathrm{raw}(t)=m_i^\mathrm{est}(t)\neq \varnothing$, we are fully confident in the original state, and thus apply a sufficiently large weight value $w_\mathrm{big}$.
    \item When $m^{\mathrm{est}}(t)$ is used due to missing or incorrect original data, the corresponding confidence value $c_i(t)$ is assigned to $w_i(t)$.
    \item If only the original state is available (\emph{i.e.}, $m^{\mathrm{raw}}(t) \neq \varnothing, m^{\mathrm{est}}(t) = \varnothing$), a small weight value $w_\mathrm{small}$ is applied to allow for flexibility.
    \item If neither $m^{\mathrm{raw}}(t)$ nor $m^{\mathrm{est}}(t)$ is available, we set $w_i(t)=0$.
\end{itemize}

In the weight assignment explained above, we set $w_{\mathrm{big}} \gg c_i \gg w_{\mathrm{small}} \gg 0$ to ensure that the weight coefficient values accurately represent the degree of certainty. To further analyze the similarity between the imputed signal states $x^\mathrm{imp}(t)=\begin{bmatrix}
    m_1^\mathrm{imp}(t),\ldots,m_N^\mathrm{imp}(t)
\end{bmatrix}$ and a feasible choice $x^\mathrm{feas}(t)=\begin{bmatrix}
    m_1^\mathrm{feas}(t),\ldots,m_N^\mathrm{feas}(t)
\end{bmatrix} \in \mathcal{X}$, we introduce two scores: the match score $s^\mathrm{match}(t)$ and the conflict score $s^\mathrm{conflict}(t)$, defined as follows:
\begin{equation}
\begin{aligned}
    s^\mathrm{match}(t,x^\mathrm{feas}(t)) &= \sum_{i=1}^N w_i(t) \cdot \mathbb{I}\left( m_i^\mathrm{imp}(t) \neq \varnothing \; \mathrm{and} \; m_i^\mathrm{imp}(t)  = m_i^\mathrm{feas}(t) \right),\\
    s^\mathrm{conflict}(t,x^\mathrm{feas}(t)) &= \sum_{i=1}^N w_i(t) \cdot \mathbb{I}\left( m_i^\mathrm{imp}(t) \neq \varnothing \; \mathrm{and} \; m_i^\mathrm{imp}(t)  \neq m_i^\mathrm{feas}(t) \right),
\end{aligned}
\end{equation}
where $\mathbb{I}(\cdot)$ is the indicator function, which equals one if the condition is satisfied and zero otherwise. 

We then determine the signal state using the following logic:
\begin{equation}
\label{eqn:min_conflict}
    x^*(t) = \argmin_{x^\mathrm{feas}(t) \in \Gamma} s^\mathrm{conflict}(t,x^\mathrm{feas}(t)),
\end{equation}
where
\begin{equation}
    \Gamma = \left\{x \mid x = \argmax_{x^\mathrm{feas}(t) \in \mathcal{X}} s^\mathrm{match}(t,x^\mathrm{feas}(t)) \right\}.
\end{equation}
Intuitively, $s^{\mathrm{match}}$ represents the sum of weight coefficients corresponding to the movements where the imputed state matches the feasible state. Conversely, $s^{\mathrm{conflict}}$ is the sum of weight coefficients for the movements where the imputed state differs from the feasible state. By selecting the state $x^*(t)$ as the feasible state $x^{\mathrm{feas}}(t)$ with the highest $s^{\mathrm{match}}$ and the lowest $s^{\mathrm{conflict}}$, we aim to ensure that movements with a higher degree of certainty, indicated by a larger $w_i(t)$, are more likely to match the final state.

Note that if the calculation in Equation~\eqref{eqn:min_conflict} yields multiple solutions, and the state at the previous time step, $x^*(t-1)$, is among these solutions, we will select this previous state as the current value, \emph{i.e.}, $x^*(t)=x^*(t-1)$. This approach ensures continuity and consistency in the signal state. With this method, we have now determined a complete signal state $x^*(t)$ at each time instance $t$, adhering to the constraints of the ring-and-barrier diagram utilized in traffic signal design.

\subsubsection{Temporal Filtering to the Signal State Sequence}

Applying the aforementioned procedure for each time instance, we obtain a complete traffic signal state sequence $\textbf{x}^* = \begin{bmatrix}
    x^*(1), \ldots, x^*(t), \ldots, x^*(T)
\end{bmatrix}$. However, this sequence may contain movements where the duration of either a green or red signal is excessively short. Typically, a green signal state should last at least three to five seconds \citep{urbanik2015signal-signal-mannual}. If a signal duration is found to be too short (excluding the initial state in the scenario), it is likely due to insufficient trajectory data or inaccuracies of estimated states. To address this issue, we apply sequential smoothing to the sequence to ensure more realistic signal durations. Specifically, we find periods where the traffic light changes too quickly. We smooth out those periods by making the light stay the same color as the previous or next interval. This process prevents quick, unrealistic changes in traffic light states and improves accuracy.

Mathematically, denote an interval \((t_{\mathrm{start}}, t_{\mathrm{end}})\) where there exists at least one movement in $\textbf{x}^*$ such that \(t_{\mathrm{end}} - t_{\mathrm{start}} \leq t_{\mathrm{min}}\), and where one of the following conditions is satisfied:
\begin{equation}
    \begin{cases}
        m_i(\tau) = G \quad \forall \tau \in [t_{\mathrm{start}}, t_{\mathrm{end}}], \quad m_i(t_{\mathrm{start}} - 1) = m_i(t_{\mathrm{end}} + 1) = R,\\
        m_i(\tau) = R \quad \forall \tau \in [t_{\mathrm{start}}, t_{\mathrm{end}}], \quad m_i(t_{\mathrm{start}} - 1) = m_i(t_{\mathrm{end}} + 1) = G.
    \end{cases}
\end{equation}
Define the set of all such time intervals as \(\mathcal{T}\). For every \((t_{j, \mathrm{start}}, t_{j, \mathrm{end}}) \in \mathcal{T}\), we smooth the sequence $\textbf{x}^*$ by updating the signal state as follows:
\begin{equation}
    x^*(\tau) := x^*(t_{j, \mathrm{start}} - 1) \quad \forall \tau \in [t_{j, \mathrm{start}}, t_{j, \mathrm{end}}].
\end{equation}
By applying this method, we ensure that the traffic signal state within each identified interval is consistent with the preceding state, thereby eliminating unreasonably short signal state periods.

After accurately determining the traffic signal sequence $\textbf{x}^*$ with only red and green signals, we now incorporate yellow lights into the sequence. A yellow signal with a fixed duration of \(t_{\mathrm{yellow}}\) is added at the end of every green signal period as follows:
\begin{equation}
    m_i^*(t) := Y \quad \forall \tau \in (t - t_{\mathrm{yellow}}, t], \; \text{if} \; m_i^*(t) = G \; \text{and} \; m_i^*(t + 1) = R.
\end{equation}

This completes the design of our methodology. It is important to note that the WOMD specifies the shape of the traffic lights, such as whether they are standard round signals or signals with arrows. It also provides traffic light information for bicycle lanes. Our updated traffic signal sequence $\textbf{x}^*$ adheres to the WOMD definitions, ensuring that the revised data can seamlessly integrate with and replace the original dataset for easy use by users. For further details, please refer to the Appendix~\ref{sec:appendix_B}.

\section{Results}

In this section, we first validate the performance of the proposed method using simulated data. Then, we present a series of case studies to illustrate the effectiveness of our proposed method in enhancing traffic signal information in the WOMD. We provide a qualitative analysis of several sample scenarios, highlighting the method's capability to handle diverse and complex real-world intersections. Subsequently, to comprehensively evaluate the performance of our method, we apply it to the entire WOMD and assess its effectiveness using two key quantitative metrics. These case studies collectively demonstrate the robustness and adaptability of our method in imputing and correcting traffic signal data.

\begin{figure}[h!]
    \centering
    \includegraphics[width=0.6\textwidth]{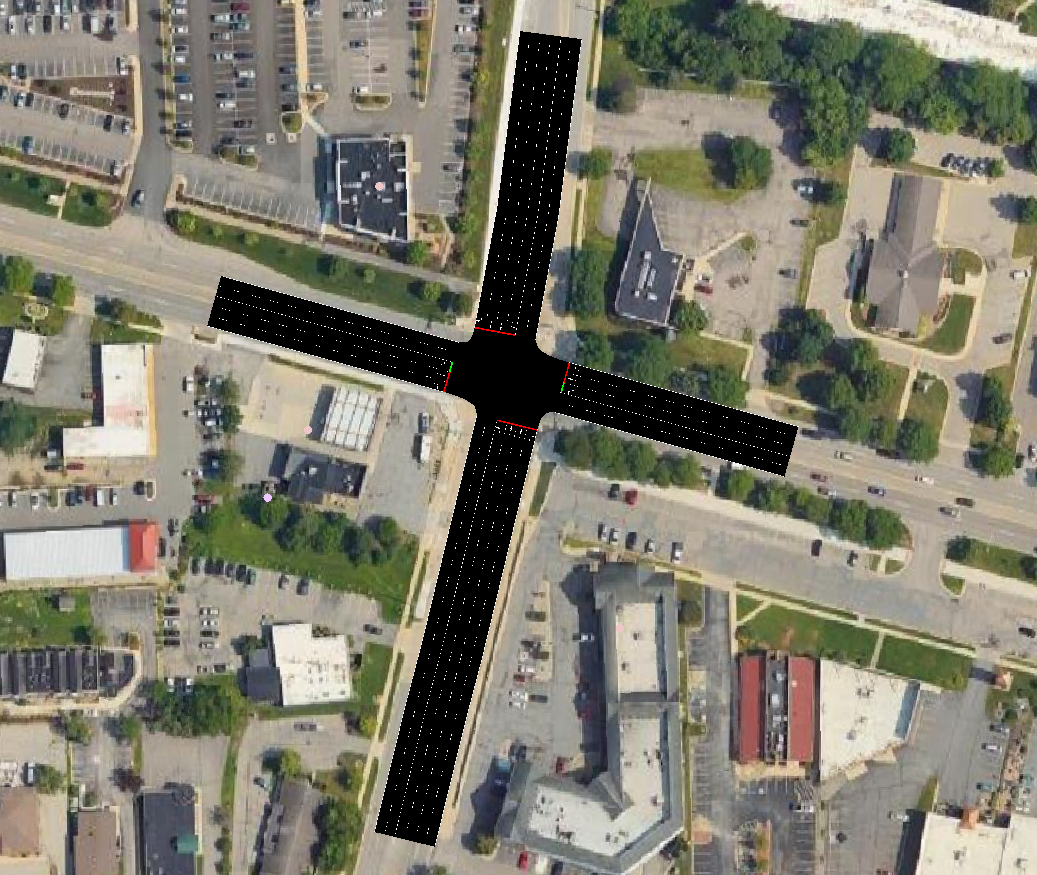}
    \vspace{2mm}
    \caption{Illustration figure of the SUMO simulated intersection (Washtenaw Avenue – S Huron Parkway) located at Ann Arbor, Michigan.}
    \label{fig:SUMO-demo}
\end{figure}


\subsection{Simulation Validation}

Before applying the proposed method on WOMD, we first evaluate its performance using simulated data, which includes ground-truth traffic signal states to quantify the estimation error. To achieve this, we built a real-world signalized intersection—Washtenaw Avenue and S Huron Parkway in Ann Arbor, Michigan—in the SUMO simulator \citep{lopez2018microscopic-sumo}, as shown in Figure~\ref{fig:SUMO-demo}. This intersection is a representative four-leg configuration with multiple through and turning lanes.

\begin{figure}
    \centering
    \includegraphics[width=0.9\textwidth]{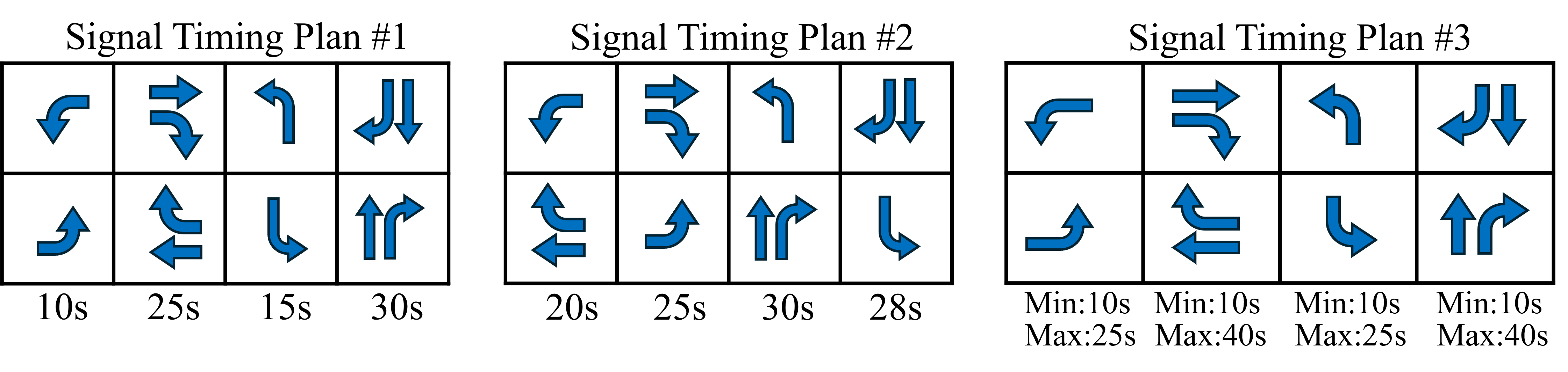}
    \caption{Diverse traffic signal settings include both static (signal timing plans \#1 and \#2) and actuated signal controls (signal timing plan \#3) for simulation validation.}
    \label{fig:signal-settings}
\end{figure}

To evaluate the robustness of our method under varying signal control strategies, we implemented three traffic signal timing plans in SUMO, including both fixed-time and actuated control schemes, as shown in Figure~\ref{fig:signal-settings}. Timing Plans \#1 and \#2 are fixed-time settings with different phase sequences and green time splits, while Timing Plan \#3 employs actuated control with a vehicle extension of 2 seconds and specified minimum and maximum green durations. All plans use a yellow phase duration of 3 seconds.

For each timing plan, we ran a one-hour simulation at 10 Hz, collecting vehicle trajectory data and ground-truth traffic signal states. To emulate the data structure of WOMD, we segmented the simulation output into 9-second intervals. Within each 9-second segment, we randomly selected one vehicle to represent the Waymo self-driving car (SDC). For the traffic lights facing the SDC, the signal states were assumed to be observable. For all other directions, we simulated missing data with a probability of 80\%, reflecting the empirical missing rate in WOMD. Additionally, observed signals were corrupted with a 5\% probability, randomly flipping their states among red, yellow, and green, thereby replicating both the missing and erroneous data characteristics of WOMD.

To evaluate performance, we define estimation accuracy as the percentage of traffic signal states correctly inferred by the model. Formally, for a signalized intersection with $N$ movements, where movement $i \in \left\{1,2,…N \right\}$ controls $L_i$ lanes, and over a total of $T$ timesteps, the total number of traffic light states is
\begin{equation}
    {M^{\text{total}} = \sum_{i=1}^{N}\sum_{t=1}^{T}L_i}.
\end{equation}
The number of correctly estimated signal states is
\begin{equation}
    M^{\text{correct}} = \sum_{i=1}^{N} \sum_{t=1}^{T} \left( L_i \cdot \mathbb{I}\left( m_i^{\text{est}}(t) = m_i^{\text{gt}}(t) \right) \right),
\end{equation}
where $\mathbb{I}(\cdot)$ is an indicator function that equals one if the estimated traffic light state $m_i^{\text{est}}(t)$ is the same as the ground-truth state $m_i^{\text{gt}}(t)$, otherwise zero. Therefore, the estimation accuracy can be calculated as
\begin{equation}
    Accuracy=\frac{M^{\text{correct}}}{M^{\text{total}}}.
\end{equation}
The results, summarized in Table~\ref{tab:signal-timing-accuracy}, show that our method achieves high estimation accuracy (exceeding 96\%) across all signal timing settings, including both fixed and actuated controls. Additionally, we observe slightly higher accuracy in fixed-time signal settings compared to the actuated case. This is expected, as the more dynamic and adaptive nature of actuated signals introduces additional variability, making accurate inference more challenging from trajectory data.

\begin{table}[htbp]
    \centering
    \begin{threeparttable}
        \caption{Estimation accuracy in simulation results.}
        \label{tab:signal-timing-accuracy}
        \begin{tabular}{lccc}
            \toprule
            \textbf{Signal Timing Plans} & \textbf{\#1} & \textbf{\#2} & \textbf{\#3} \\
            \midrule
            Accuracy & 96.98\% & 97.72\% & 96.08\% \\
            \bottomrule
        \end{tabular}
    \end{threeparttable}
\end{table}

We also use the simulation environment to calibrate certain parameters of the proposed method. The candidate value sets for each parameter are as follows: $a_{\mathrm{green}}\in[0.5,2.0]$, $a_{\mathrm{red}}\in[-2.0,-0.5]$, $v_{\mathrm{green}}\in[3.0,5.0]$, $v_{\mathrm{red}}\in[0.5,2.0]$, all with 0.5 as the resolution, and $\theta\in[0.8,1.0,1.2]$. This results in 960 unique parameter combinations. For each combination, we evaluated the overall estimation accuracy across all three traffic signal timing plans in the SUMO environment. The best-performing parameter set was: $a_{\mathrm{green}}=0.5$, $a_{\mathrm{red}}=-2.0$, $v_{\mathrm{green}}=3.5$, $v_{\mathrm{red}}=0.5$, and $\theta=1.0$. Therefore, the  Table~\ref{table:parameters-setup} lists all parameter setup for this study used in the following WOMD experiments.

\begin{table}[h!]
    \centering
    \begin{threeparttable}
    \begin{tabular}{ccc}
        \toprule
        Parameter & Meaning & Value \\
        \midrule
        $T$ & Total number of discrete time instances in one scenario & 91\\ 
        $\Delta t$ & Past and future time horizon & 10\\ 
        
        $a_{\mathrm{green}} (\mathrm{m/s^2})$ & Green light threshold for weighted mean acceleration & 0.5\\
        $a_{\mathrm{red}} (\mathrm{m/s^2})$ &  Red light threshold for weighted mean acceleration & -2.0\\
        $v_{\mathrm{green}} (\mathrm{m/s})$ &  Green light threshold for weighted mean velocity & 3.5\\
        $v_{\mathrm{red}} (\mathrm{m/s})$ &  Red light threshold for weighted mean velocity & 0.5\\
        $\theta$ & Threshold for estimated movement confidence value & 1.0\\
        $w_{\mathrm{big}}$ & Sufficiently large weight coefficient for imputed signal state & 100\\
        $w_{\mathrm{small}}$ & Sufficiently small weight coefficient for imputed signal state & 0.1\\
        $t_{\mathrm{min}}$ & Minimum interval for temporal filtering & 30\\
        $t_{\mathrm{yellow}}$ & Fixed duration for yellow signal & 20\\
        \bottomrule
    \end{tabular}
    \caption{Parameters setup.}
    \label{table:parameters-setup}
    \end{threeparttable}
\end{table}

\subsection{Scenario Analysis}

 In this subsection, we provide a qualitative analysis of several WOMD scenarios, demonstrating the method's capability to handle diverse and complex real-world intersections.

\begin{figure}[t]
    \centering
    \begin{subfigure}[b]{1\textwidth}
        \centering
        \includegraphics[width=\textwidth]{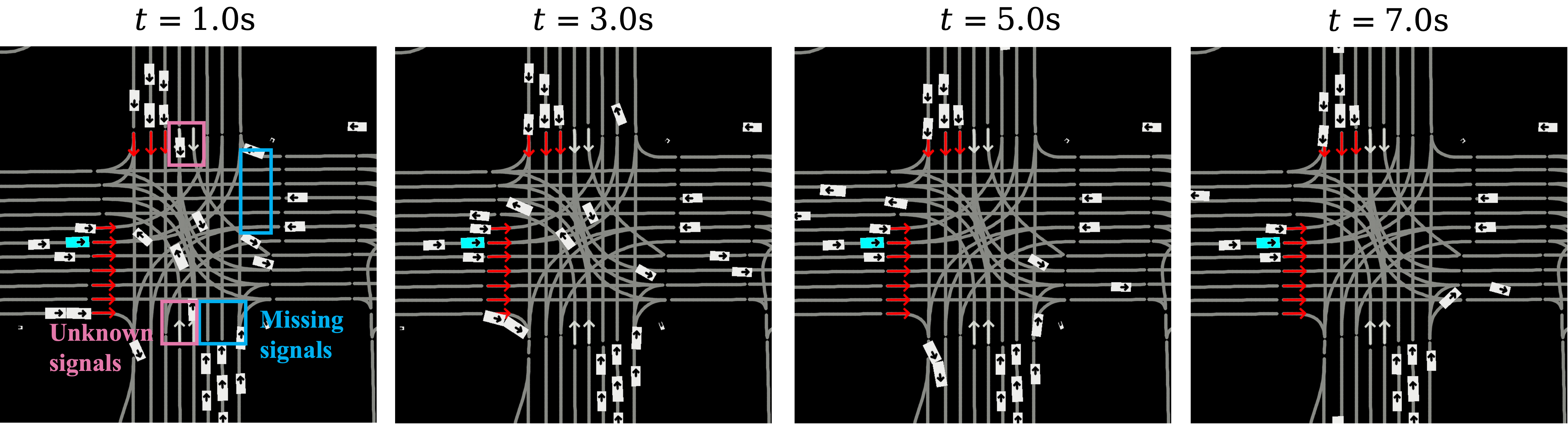} 
        \caption{}
        \label{fig:case1-raw}
    \end{subfigure}
    \begin{subfigure}[b]{1\textwidth}
        \centering
        \includegraphics[width=\textwidth]{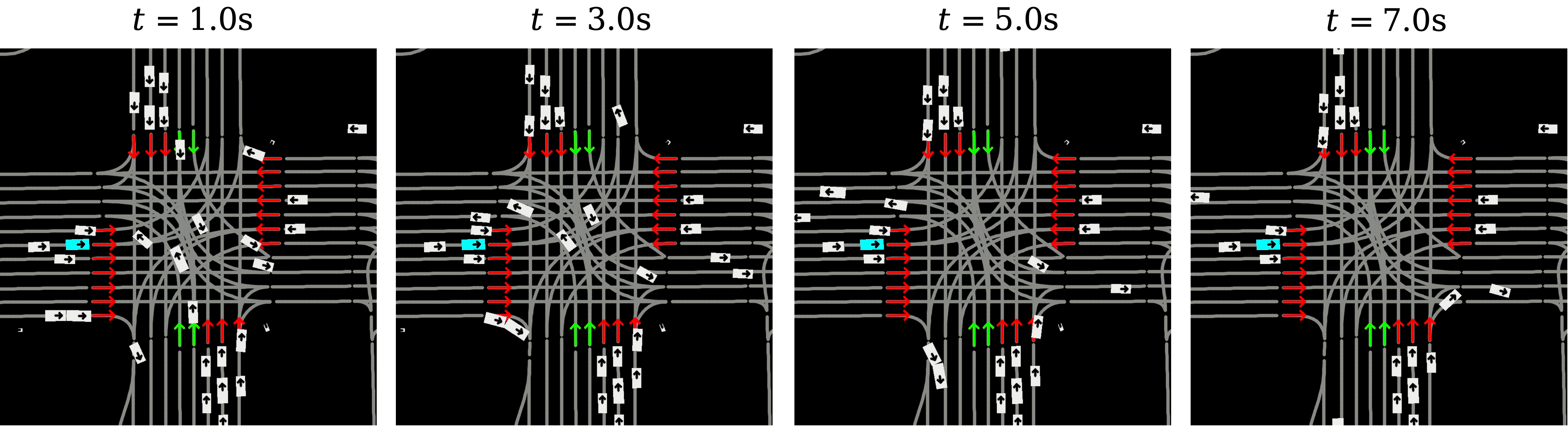}
        \caption{}
        \label{fig:case1-ours}
    \end{subfigure}
    \vspace{2mm}
    \caption{Demonstration of scenario 1. (a) Original data from the WOMD, (b) Data after applying the proposed method.}
    \label{fig:case1}
\end{figure}

\subsubsection{Scenario 1: Imputation of Unknown and Missing Traffic Signal Information}

In the first scenario (ID: 914f16ffef6529c5), we examine a typical four-way signalized intersection, as illustrated in Figure~\ref{fig:case1}. The raw data, comprising trajectory, map, and traffic signal information, is depicted in Figure~\ref{fig:case1-raw}. For clarity, we present four moments within the 9-second data segment. In the original data, only three out of eight movements—eastbound left turn, eastbound through, and southbound through-are provided. While the positions for the traffic signals controlling the left-turn movements on the southbound and northbound approaches are included, their states are unknown. Both the positions and states of the traffic signals for the entire westbound approach are missing.

Figure~\ref{fig:case1-ours} shows the results after applying our proposed method. The method successfully identifies all missing traffic signal positions, creating a complete traffic signal layout for the intersection. Furthermore, it accurately estimates the traffic signal states for each movement, providing comprehensive traffic signal data for this scenario. We can evaluate the correctness of the estimated traffic signal states using the vehicle trajectories. As observed in the figures, vehicles are making left turns in both the northbound and southbound approaches, while vehicles in all other non-right-turn movements remain stationary. This observation is consistent with our estimated traffic light states, demonstrating the effectiveness of our method.

\begin{figure}[t]
    \centering
    \begin{subfigure}[b]{1\textwidth}
        \centering
        \includegraphics[width=\textwidth]{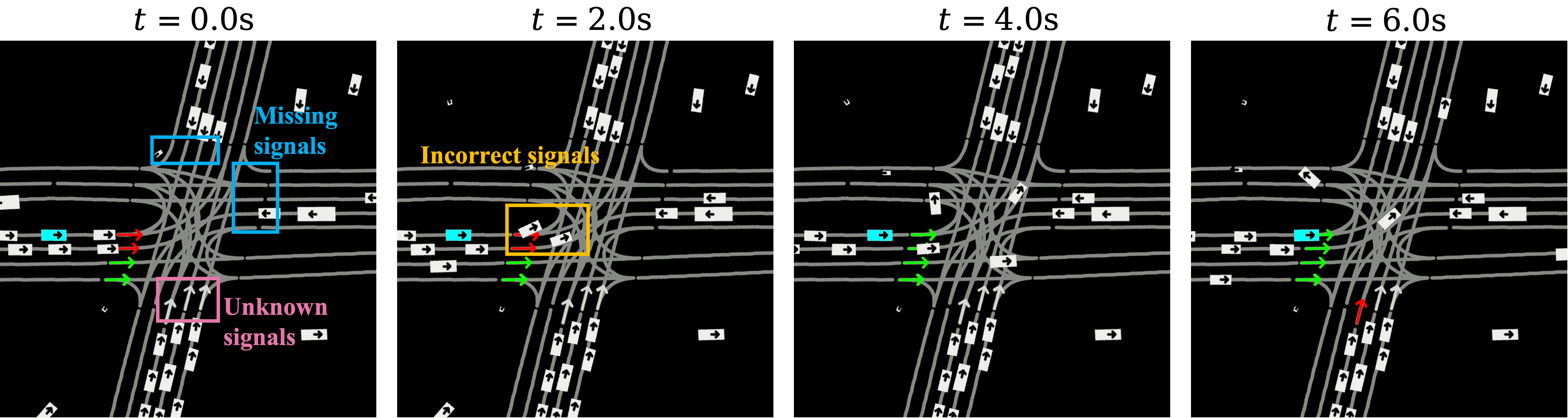} 
        \caption{}
        \label{fig:case2-raw}
    \end{subfigure}
    \begin{subfigure}[b]{1\textwidth}
        \centering
        \includegraphics[width=\textwidth]{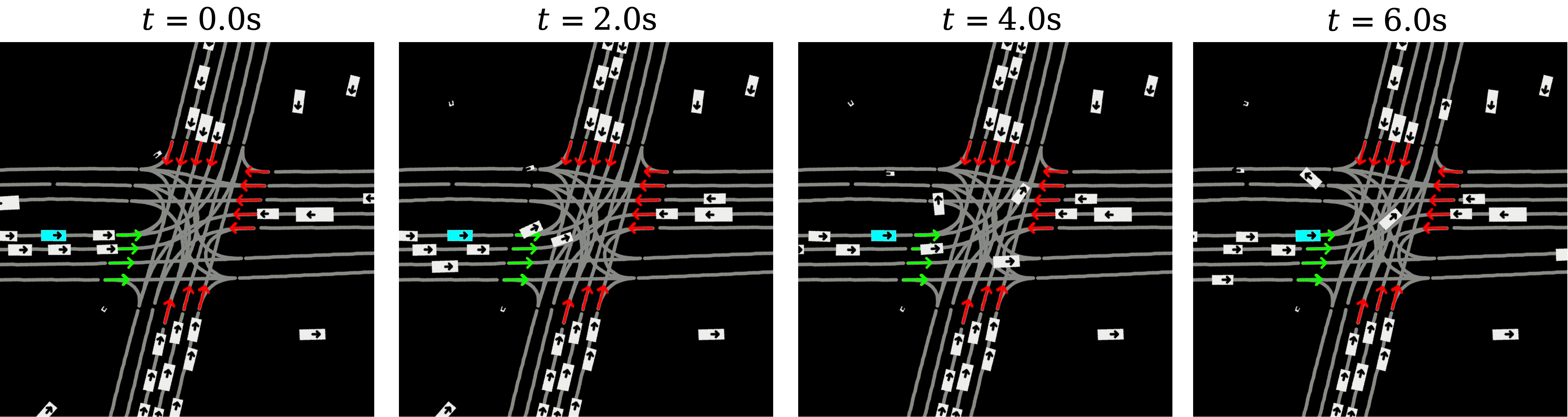}
        \caption{}
        \label{fig:case2-ours}
    \end{subfigure}
    \vspace{2mm}
    \caption{Demonstration of scenario 2. (a) Original data from the WOMD, (b) Data after applying the proposed method.}
    \label{fig:case2}
\end{figure}

\subsubsection{Scenario 2: Rectification of Inaccurate Traffic Signal Information}

In the second scenario (ID: 2d84b1ab55ab81d3), the raw data not only contains unknown and missing traffic signals but also incorrect signal state records. At $t=2.0 \text{s}$, as seen in the second image of Figure \ref{fig:case2-raw}, several vehicles from the west are passing the stop bar at high velocity, indicating a green light for the corresponding movement. However, the raw data incorrectly records the signal as red during this period. Our method successfully corrects this error, as shown in Figure~\ref{fig:case2-ours}, thereby enhancing the data quality and ensuring a cleaner dataset.

\begin{figure}[t]
    \centering
    \begin{subfigure}[b]{1\textwidth}
        \centering
        \includegraphics[width=\textwidth]{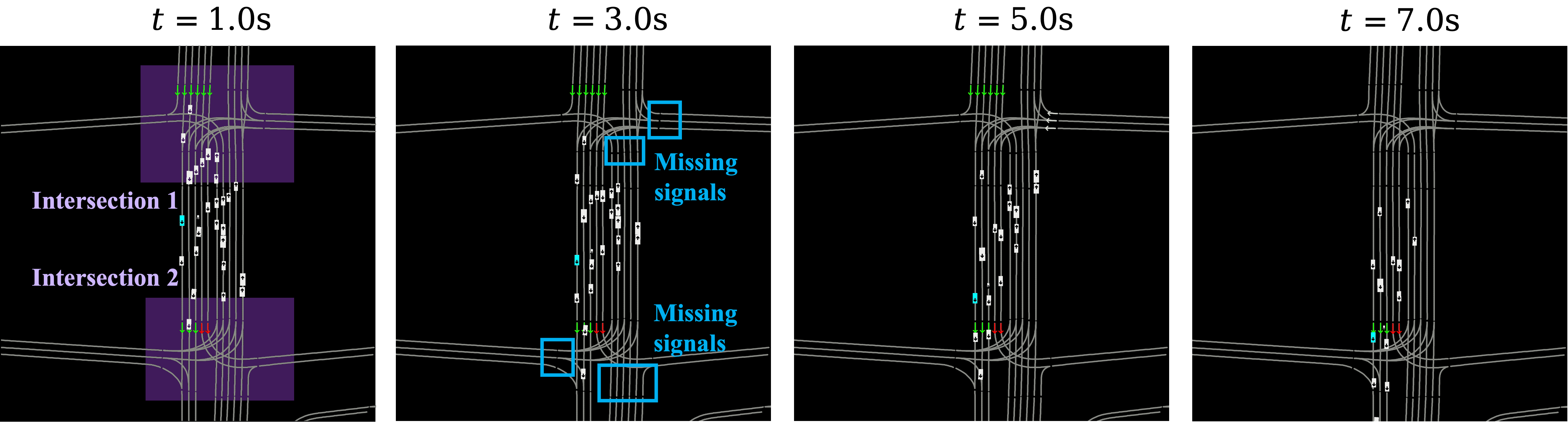} 
        \caption{}
        \label{fig:case3-raw}
    \end{subfigure}
    \begin{subfigure}[b]{1\textwidth}
        \centering
        \includegraphics[width=\textwidth]{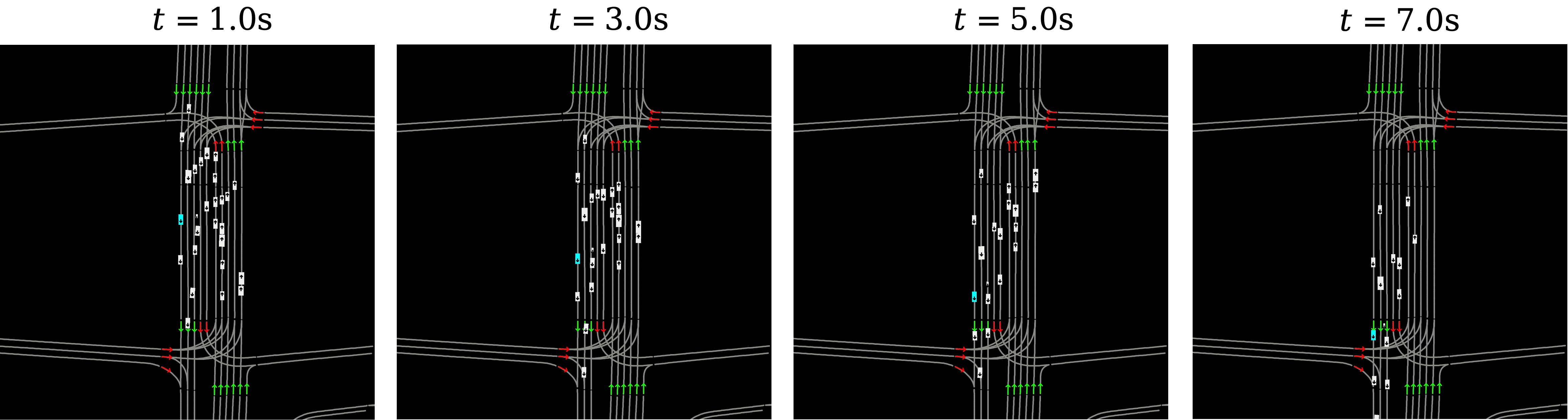}
        \caption{}
        \label{fig:case3-ours}
    \end{subfigure}
    \vspace{2mm}
    \caption{Demonstration of scenario 3. (a) Original data from the WOMD, (b) Data after applying the proposed method.}
    \label{fig:case3}
\end{figure}

\subsubsection{Scenario 3: Adaptability to Diverse Intersection Configurations}

\begin{figure}[p!]
    \centering
    \includegraphics[width=0.85\textwidth]{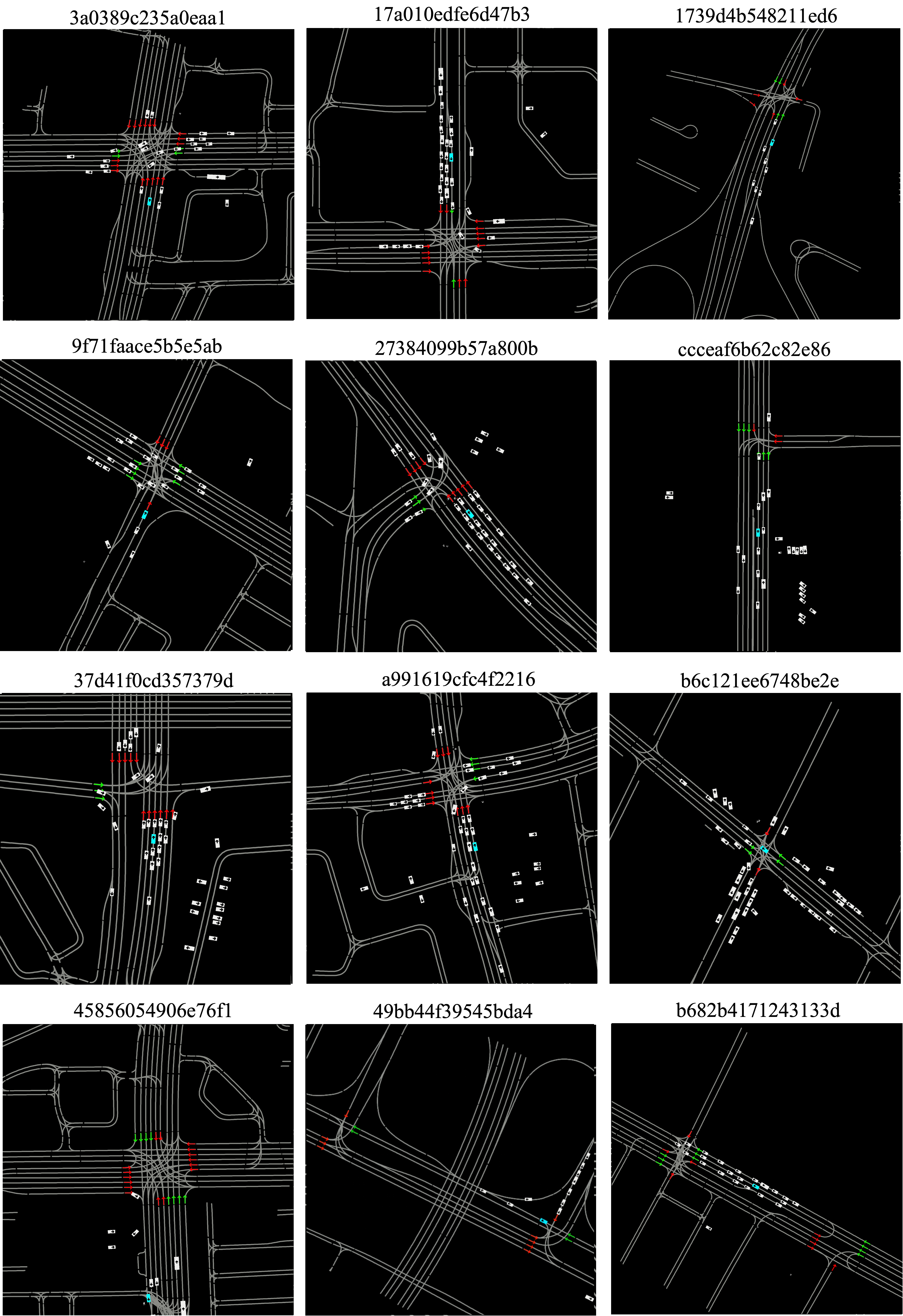} 
    \vspace{2mm}
    \caption{Demonstration of the proposed method in diverse real-world intersections. The title for each sub-figure is the corresponding scenario ID in the WOMD.}
    \label{fig:matrix}
\end{figure}

The proposed method demonstrates adaptability to various intersection configurations, including complex and irregular layouts encountered in real-world scenarios. In the third scenario (ID: 851e351605bc2291), illustrated in Figure~\ref{fig:case3-raw}, two signalized intersections display several intricate features: the intersections are closely spaced, some approaches consist of one-way roads, and not all traffic movements are included. Despite these complexities, our method successfully provides reasonable traffic signal states for both intersections that align with the observed vehicle trajectories, as shown in Figure~\ref{fig:case3-ours}. This example highlights the robustness and flexibility of our method in handling diverse intersection configurations. More results of the proposed method in various scenarios are illustrated in Figure~\ref{fig:matrix}.

\subsection{Quantitative Evaluation Using the Entire WOMD}

To systematically validate the performance of our proposed method, we applied it to the entire WOMD, encompassing both the training and validation splits, but excluding the test split as it contains only 1-second trajectories per scenario, which is insufficient for our estimation. The complete dataset comprises over 530,000 scenarios, of which more than 360,000 are relevant scenarios featuring signalized intersections. To quantify the effectiveness of our method in imputing and correcting traffic signal data, we employed two metrics. These metrics allow us to objectively assess and demonstrate the improvements that our method introduces to the WOMD.

\subsubsection{Traffic Signal imputation Rate}
The traffic signal imputation rate measures the percentage of traffic signal states provided by our method out of the total traffic signal states in a scenario. For example, if the raw WOMD data provides traffic signal states for only one approach in a four-way intersection (as in $t=0$ seconds in Scenario 2 shown in Figure~\ref{fig:case2-raw}), the imputation rate would be approximately $75\%$. This metric highlights the severity of data unknown and missing issues and underscores the importance of our method in providing complete traffic signal information.

The imputation rate is formally defined as follows. Given a signalized intersection with $N$ movements, assume each movement $i \in \left\{ 1, 2, \cdots N\right\}$ controls $L_i$ lanes, where $L_i \in \left\{1, 2, 3,\cdots \right\}$, then the total number of traffic light states for this scenario $S$ is
\begin{equation}
    {M_{S}^{\text{total}} = \sum_{i=1}^{N}\sum_{t=1}^{T}L_i}.
\end{equation}
The number of traffic light states provided by the proposed method is
\begin{equation}
    M_{S}^{\text{impute}} = \sum_{i=1}^{N} \sum_{t=1}^{T} \left( L_i \cdot \mathbb{I}\left( m_i^{\text{raw}}(t) = \varnothing \right) \right),
\end{equation}
where $\mathbb{I}(\cdot)$ is the indicator function that equals one if $m_i^{\text{raw}}(t) = \varnothing$ and zero otherwise. This function indicates whether a movement's signal state was provided by the original data or filled in through imputation. For a set of scenarios $\mathcal{S}$ with signalized intersections, the traffic signal imputation rate is
\begin{equation}
    {\eta_{\mathcal{S}} = \frac{\sum_{S\in \mathcal{S}}M_{S}^{\text{impute}}}{\sum_{S\in \mathcal{S}}{M_{S}^\text{total}}}}.
\end{equation}

Our method achieves a 71.7\% traffic signal imputation rate. This implies that our method successfully fills the gaps for approximately three-fourths of the total traffic signal states, significantly enhancing the completeness of the original dataset.

\subsubsection{Red Light Violation Rate}

To validate the accuracy of our method in the absence of ground-truth traffic signal states, we calculate the red light violation rate based on observed vehicle trajectories. The rationale is that red light running is a rare event in reality. Thus, if our estimated traffic light states are accurate, there should be a limited number of red light running events. This metric effectively evaluates the performance of our estimated traffic signals and the data correction capabilities for erroneous data in the WOMD.

We define a red light running violation as an event in which a vehicle crosses the stop bar while the corresponding traffic signal is red. A scenario is classified as a red light running scenario if at least one such violation occurs during the 9-second time window. The red light violation rate $\xi$ is then computed as the ratio of red light running scenarios to the total number of scenarios involving signalized intersections.

We computed this metric for both the original and our enhanced traffic signal data. In the original data, the red light violation rate is $\xi^\mathrm{raw}=15.7\%$, which is unreasonably high compared to real-world situations and indicates data inaccuracies. With our enhanced traffic signal information, this rate decreases to $\xi^\mathrm{ours} = 2.9\%$, highlighting the superior performance of our method. Note that when calculating the red light violation rate of the proposed method, we consider all movements, including those missing or unknown in the original dataset. This result demonstrates the consistency between the enhanced traffic signal information and the observed trajectory data, validating the effectiveness of our model and its ability to improve data quality.

\begin{figure}
    \centering
    \begin{minipage}[t]{0.49\textwidth}
        \centering
        \includegraphics[width=\linewidth]{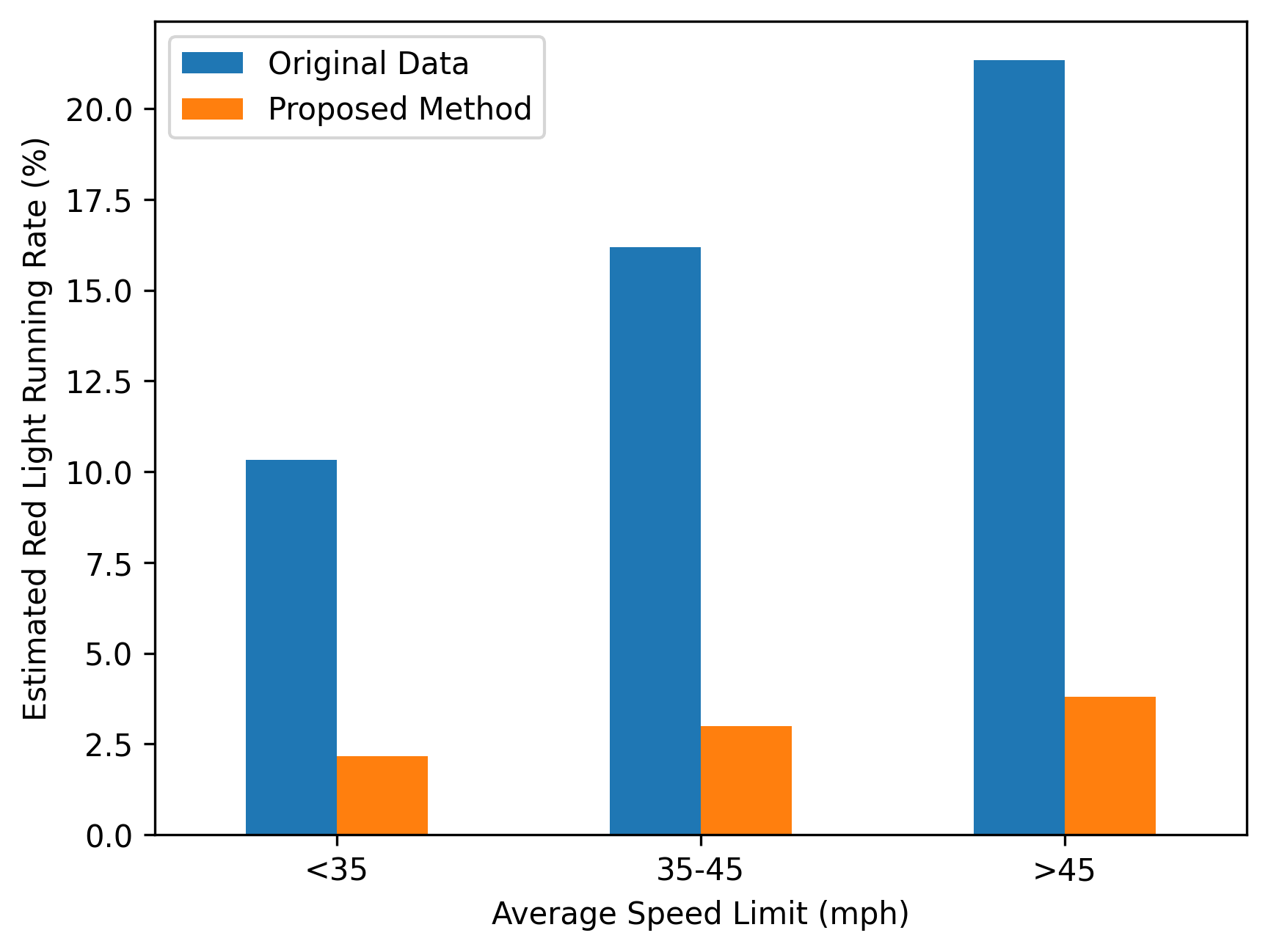}
        \caption{Estimated red-light running rates by average intersection speed limit category.}
        \label{fig:average-speed-limit}
    \end{minipage}
    \hfill
    \begin{minipage}[t]{0.49\textwidth}
        \centering
        \includegraphics[width=\linewidth]{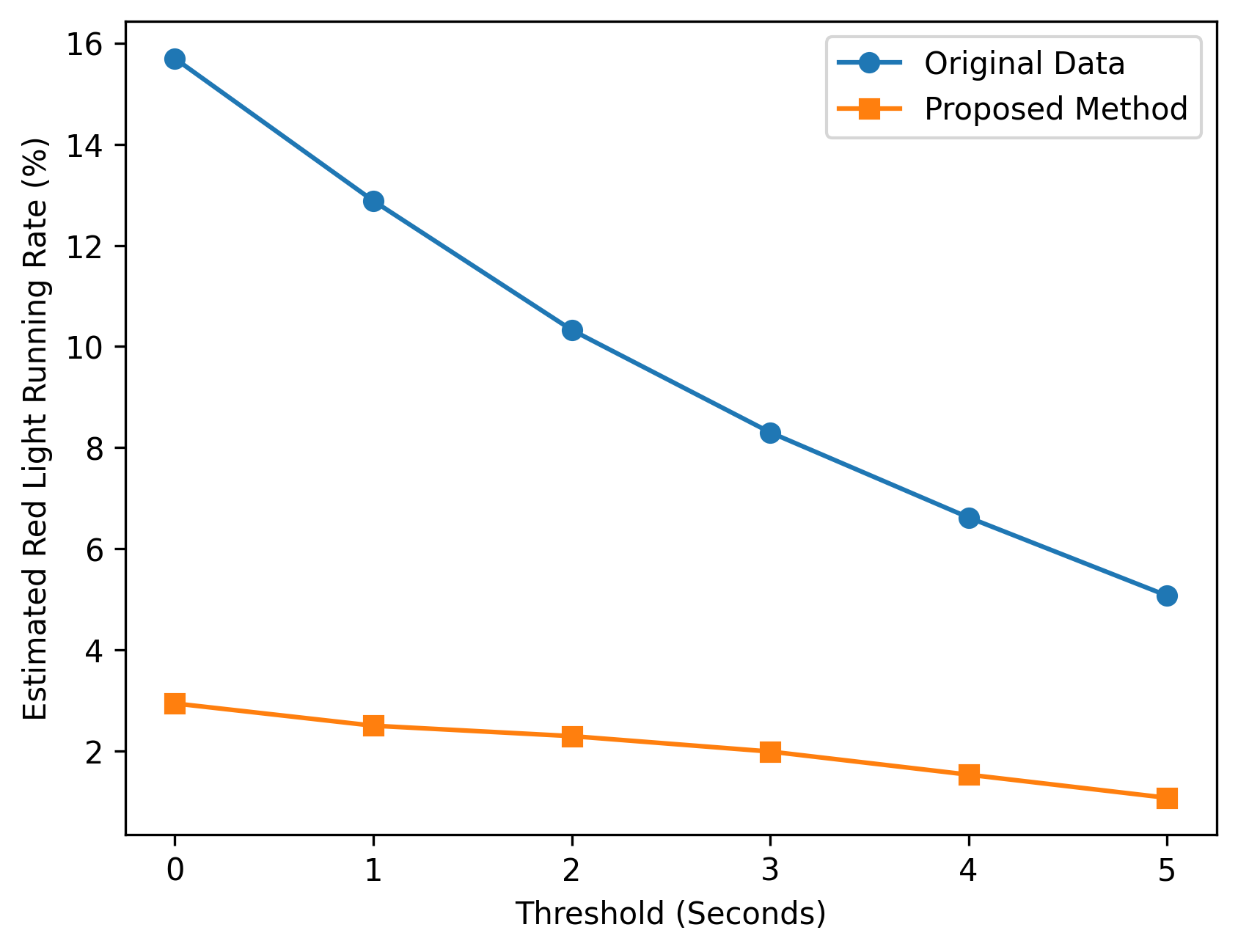}
        \caption{Estimated red-light running rates under varying red phase thresholds.}
        \label{fig:sensitivity}
    \end{minipage}
\end{figure}

To further analyze the results, we classified scenarios according to the average speed limits of all the approaching lanes at each intersection, which serves as a proxy for the type of road and the environmental complexity. Specifically, the scenarios were grouped into three categories: below 35 mph, 35-45 mph, and above 45 mph. The estimated red-light running rates, both from the raw data and after applying our proposed method, are presented in Figure~\ref{fig:average-speed-limit}. We observe that lower-speed scenarios tend to exhibit lower estimated red-light running rates in both the raw and estimated data. One possible explanation is that the Waymo SDC spends more time navigating low-speed intersections, leading to more agent trajectory observations and thus richer data for signal state estimation. Lower speeds also likely reduce sensor noise, improving data quality. Additionally, longer durations within the intersection allow the perception system to more clearly recognize traffic signals, which may account for the higher quality of the original signal data in these cases.

Furthermore, the red light running violations depend on how strictly they are defined. Events that occur shortly after the signal turns red are more common, while those occurring well into the red phase are rarer. To account for this, we conducted a sensitivity analysis by introducing a threshold on how long the signal must have been red before a vehicle’s passage is considered a violation. Specifically, we vary the threshold from 0 to 5 seconds. For example, under a 3-second threshold, a vehicle is only counted as running the red light if it crosses the stop bar at least 3 seconds after the signal turned red. The results are shown in Figure~\ref{fig:sensitivity}. As expected, the estimated red-light running rates for both the original data and our estimated data decrease as the threshold increases. Importantly, our method consistently yields lower estimated red light running rates than the original data across all threshold settings, highlighting its effectiveness in imputing and correcting traffic signal states in WOMD.
\section{Conclusions}

Ensuring data quality is paramount in autonomous driving and AI research, as many modern methods depend heavily on machine learning algorithms, which necessitate clean and accurate datasets. In this study, we tackled the problem of missing and inaccurate traffic signal data within the WOMD. We developed a fully automated approach that leverages trajectory data and transportation domain knowledge to effectively impute and correct traffic signal information. Our method was applied across the entire WOMD, covering over 360,000 relevant scenarios involving traffic signals out of a total of 530,000 scenarios. The results demonstrated an imputation rate of $71.7\%$ and a reduction in the red-light violation rate from $15.7\%$ to $2.9\%$, underscoring the method's effectiveness and accuracy. This enhancement will improve the quality of the WOMD and benefit all research and applications built upon this dataset.

Despite these promising results, there are several areas for potential improvement. The proposed method performs well for three-way and four-way intersections that follow NEMA standards with Ring-and-Barrier structure, which represent the majority of signalized intersections in the US. However, its performance may degrade at nonstandard locations, such as intersections with five or more approaches, unusual geometries, or complex lane configurations. For example, large intersections or those with right turn lanes that are significantly separated from through-movement lanes pose challenges to reliably identify the intersection structure and correctly assign vehicle trajectories for estimation. These complexities can hinder accurate signal state estimation. Addressing these limitations could further improve the WOMD quality and contribute to the development of safer and more reliable autonomous driving systems.





\section{Acknowledgments}
This research was partially funded by the National Science Foundation (CMMI $\#2223517$). The visualization tools of the WOMD are developed based on https://github.com/zhejz/TrafficBots. Any opinions, findings, and conclusions or recommendations expressed in this material are those of the authors and do not necessarily reflect the official policy or position.

\bibliographystyle{elsarticle-harv}
\bibliography{ref.bib}

\newpage

\appendix

\section*{Appendix}

\renewcommand{\thesubsection}{\Alph{subsection}}

\subsection{Vehicle Confidence Functions}
\label{sec:appendix_A}

\begin{figure}[h]
    \centering
    \begin{subfigure}[b]{0.45\textwidth}
        \centering
        \includegraphics[width=\textwidth]{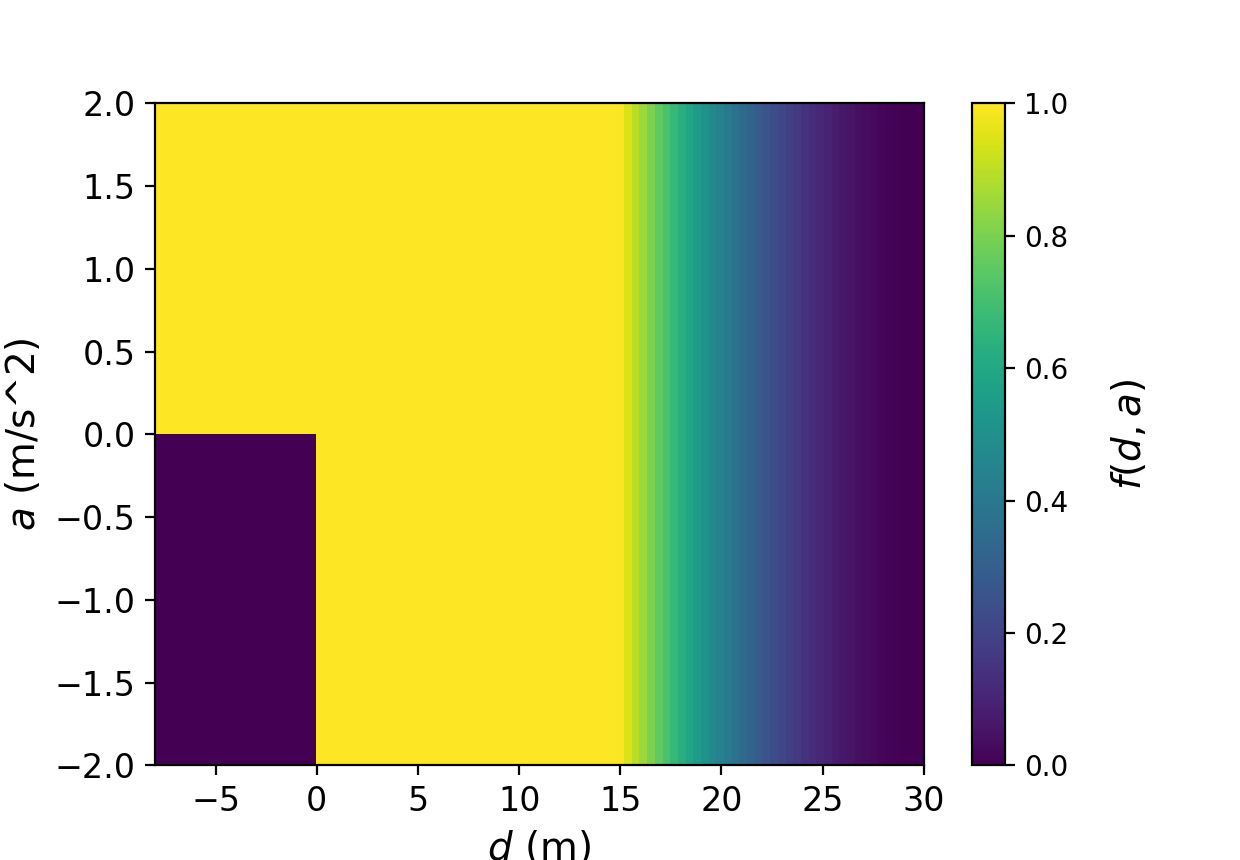}
        \caption{}
        \label{fig:heatmap_f}
    \end{subfigure}
    \hspace{5mm}
    \begin{subfigure}[b]{0.45\textwidth}
        \centering
        \includegraphics[width=\textwidth]{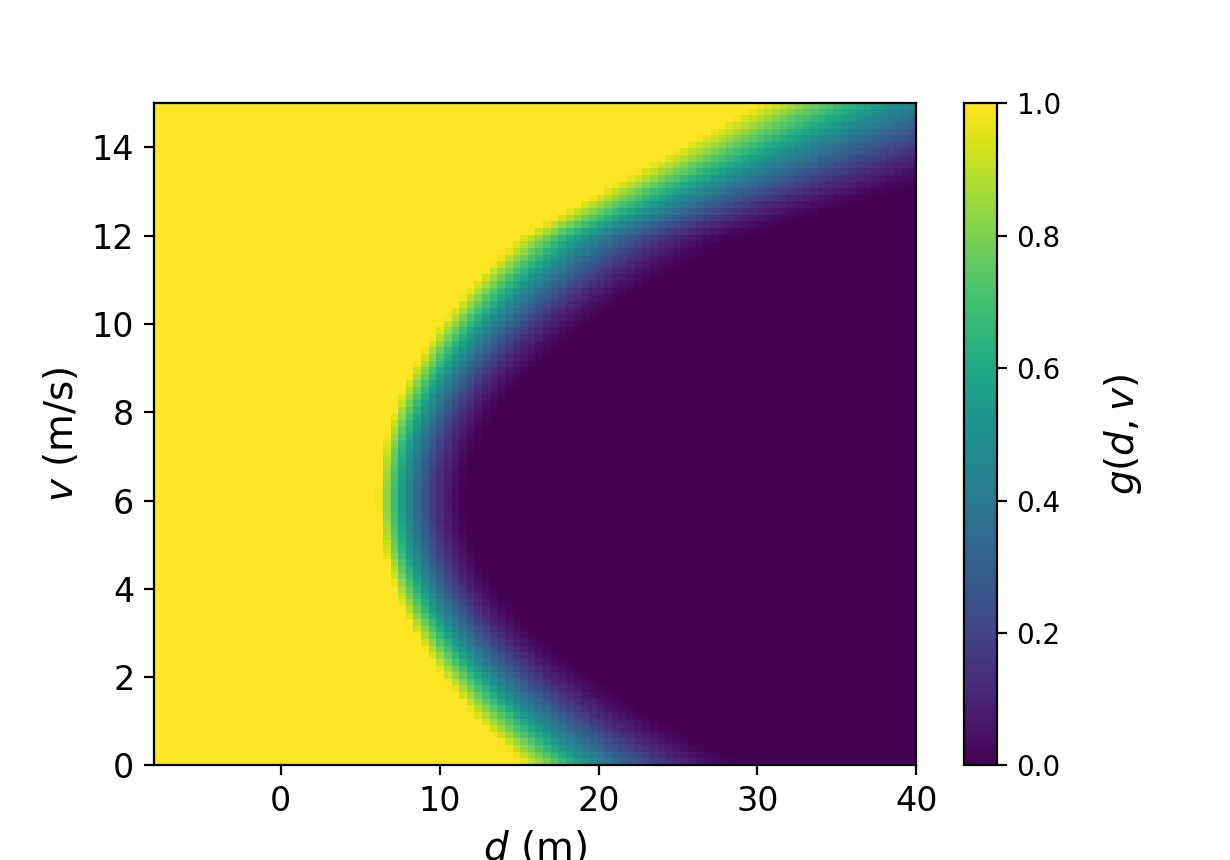}
        \caption{}
        \label{fig:heatmap_g}
    \end{subfigure}
    \caption{Heatmaps of the vehicle confidence functions. (a) Heatmap of $f$, (b) Heatmap of $g$.}
    \label{fig:heatmap}
\end{figure}

Given a vehicle at time $t$, let $d\in \mathbb{R}$ (unit: $ \mathrm{m}$) be its distance to the stopping line, where $d<0$ means the vehicle has passed the stopping line; $a \in \mathbb{R}$ (unit: $\mathrm{m/s^2}$) be its current acceleration; and $v\geq0$ (unit: $\mathrm{m/s}$) be its velocity. Those vehicles with $d\geq-8$ m are taken into consideration. Then the acceleration confidence function $f$ and the velocity confidence function $g$ are given by: ($t$ is neglected for concise expression)
\begin{equation}
\label{eqn:confidence_velocity}
f(d,a) =\begin{cases}
0, &\quad \mathrm{if}\;d<0, a<0\;\mathrm{or}\;a\geq0\;\mathrm{or}\;d>30;\\
\frac{(d-30)^2}{15^2}, &\quad \mathrm{if}\; 15<d\leq30;\\
1, &\quad \mathrm{if}\; d>30.\\
\end{cases}
\end{equation}

\begin{equation}
\label{eqn:confidence_acceleration}
g(d,v) = \begin{cases}
1, &\quad \mathrm{if}\; -8 \leq d \leq g_0(v);\\
\frac{(d-2g_0(v))^2}{(g_0(v))^2}, &\quad \mathrm{if}\; g_0(v)<d\leq 2g_0(v);\\
0, &\quad \mathrm{if}\; d>2g_0(v),
\end{cases}
\end{equation}
where
\begin{equation}
    g_0(v)=\begin{cases}
3(v-6)^2/4+6, &\quad \mathrm{if}\;v \leq 12;\\
\min\left(5(v-12)+15, 30\right), &\quad \mathrm{if}\; v>12.
\end{cases}
\end{equation}
The cutoff distance in $f$ is set as $30$ meters based on the dilemma zone boundaries provided by the Federal Highway Administration (FHWA) \citep{gordon1996traffic-dilemma-zone}. Figure \ref{fig:heatmap} shows the profile of $f$ of $g$ at different setups of distance $d$ and acceleration $a$ or velocity $v$.

\subsection{Details on the WOMD Traffic Signal Data}
\label{sec:appendix_B}

The WOMD defines eight traffic signal states: RED, YELLOW, GREEN, RED-ARROW, YELLOW-ARROW, GREEN-ARROW, RED-FLASHING, and YELLOW-FLASHING. Specifically, it differentiates between standard round signals and arrow signals, and also defines two types of flashing signals. 

In our implementation, a red or yellow signal at any movement \(i\) at any time \(t\) will be classified as an arrow signal only if the following two conditions are met:
\begin{itemize}
    \item The lane is a dedicated left-turn lane.
    \item Arrow red or yellow signals appeared in this movement in the original data of this scenario.
\end{itemize}
If these conditions are not met, the signal remains a standard round red or yellow signal.

A green signal at any movement \(i\) at any time \(t\) will be classified as an arrow green signal only if all of the following three conditions are satisfied:
\begin{itemize}
    \item The lane is a dedicated left-turn lane.
    \item Arrow green signals appeared in this movement in the original data of this scenario.
    \item At that moment, the movement is not an unprotected left turn, \emph{i.e.}, the through movement in the opposite approach is not allowed at the same time.
\end{itemize}
If any of these conditions are not met, the signal is classified as a standard round green signal.Due to the rarity of flashing signals, our method does not generate them. 

Following the WOMD definitions, our traffic light information $\textbf{x}^*$ will be applied to all lanes associated with each movement, encompassing both vehicular and bicycle lanes. Specifically, the traffic light information for bicycle lanes will be aligned with that of the nearest vehicular lane.

\end{document}